\documentclass[onluarkali,ingilizce,lisans,karton,ehb,tarihli,onsozacik,ozetacik,appendix]{itutez}

\usepackage[table,xcdraw]{xcolor}
\usepackage{amsmath}

\usepackage{lipsum}
\usepackage{graphicx}
\usepackage{afterpage}
\usepackage{lscape}
\usepackage{float}
\usepackage{mathtools}
\usepackage{listings}
\usepackage[final]{pdfpages}
\usepackage{hyperref}
\usepackage{url}
\usepackage{listings}
\usepackage{amssymb}

\usepackage{pdflscape}

\usepackage{booktabs}

\hypersetup{
    colorlinks,
    linkcolor={black!50!black},
    citecolor={black!50!black},
    urlcolor={black!80!black}
}

\definecolor{bg}{rgb}{0.95,0.95,0.95}

\yazar{Temp}{Author}{150230423}
\authorSecond{Oğuzhan}{KÖSE}{(040160457)}
\authorFirst{}{}{}
\authorThird{}{}{}

\unvan{}
 
\anabilimdali{}{Department of Control and Automation Engineering}
\programi{}{Kontrol ve Otomasyon Mühendisliği Bölümü}

\tarih{HAZİRAN, 2021}{JULY, 2021}
\tarihKucuk{HAZİRAN, 2021}{JULY, 2021}

\tezyoneticisi{Assoc. Prof. Dr. Tufan KUMBASAR}{Doç. Dr. Tufan KUMBASAR}  
\esdanismani{}{}

\baslik{REAL-WORLD APPLICATION OF VARIOUS} {TRAJECTORY PLANNING ALGORITHMS}{ON MIT RACECAR}
\title{YÖRÜNGE PLANLAMA ALGORİTMALARININ} {MIT RACECAR ÜZERİNDE GERÇEKLENMESİ}{}

\usepackage{color}
\usepackage{times}
\usepackage{amssymb,amsmath,mathptmx,amsbsy,bm}
\usepackage{caption}            
\usepackage{graphics}
\usepackage{wrapfig}
\usepackage{epsfig}
\usepackage{enumerate}
\usepackage{rotating}
\usepackage{multirow}
\usepackage{subfigure}
\usepackage{colortbl}
\usepackage{pstricks}
\usepackage{pst-plot}
\usepackage{cite}%
\usepackage{latexsym}           %
\usepackage{rotating}
\usepackage{fixltx2e} 

\def\be{\begin{equation}} %
\def\ee{\end{equation}}%
\def\beq{\begin{eqnarray}}%
\def\eeq{\end{eqnarray}}%
\def\bse{\begin{subequations}}%
\def\ese{\end{subequations}}%
\def\[{\left[}
\def\]{\right]}
\def\({\left(}
\def\){\right)}

\ithaf{To my dearest friends and family,}

\kisaltmalistesi{\hspace{-3mm}
\begin{tabular}{p{2cm}l}

{\bf RACECAR} & {\bf:} Rapid Autonomous Complex-Environment Competing Ackermann-steering Robot\\
{\bf ROS} & {\bf:} Robot Operating System\\
{\bf DWA} & {\bf:} Dynamic Window Approach\\
{\bf TEB} & {\bf:} Timed-Elastic Band\\
{\bf APF} & {\bf:} Artificial Potential Field\\
{\bf ADAS} & {\bf:} Advanced Driving Assistance System\\
{\bf RRT} & {\bf:} Randomly Exploring Random Tree\\
{\bf RRT*} & {\bf:} Randomly Exploring Random Tree Star\\
{\bf PRM} & {\bf:} Probabilistic Roadmap\\
{\bf UGV} & {\bf:} Unmanned Ground Vehicle\\
{\bf GPU} & {\bf:} Graphical Processing Unit\\
{\bf CPU} & {\bf:} Central Processing Unit\\
{\bf RAM} & {\bf:} Random Access Memory\\
{\bf FOV} & {\bf:} Field of View\\
{\bf FPS} & {\bf:} Frame per Second\\
{\bf IMU} & {\bf:} Inertial Measurement Unit\\
{\bf VESC} & {\bf:} Vedder Electronic Speed Controller\\
{\bf BLDC} & {\bf:} Brushless Direct Current Motor\\
{\bf XML} & {\bf:} Extensible Markup Language\\
{\bf RPC} & {\bf:} Remote Procedure Call\\
{\bf AI2S} & {\bf:} Artificial Intelligence and Intelligent Systems Laboratory\\

\end{tabular}

}
\sembollistesi{\hspace{-3mm}
\begin{tabular}{p{2cm}l}

{$\gamma$} & {\bf:} Steering Angle\\
{$L$} & {\bf:} Wheelbase\\
{$R$} & {\bf:} Turning Radius of the Vehicle\\

\end{tabular}

}
\onsoz{I would like to thank my advisor Assoc. Prof. Tufan KUMBASAR for his support and guidance throughout the graduation project period. I am grateful that Artificial Intelligence and Intelligent Systems (AI2S) Laboratory opened its doors and provided a healthy study environment for this project. This research is supported by the project (118E807) of Scientific and Technological Research Council of Turkey.

I also thank my beloved family, dearest friends and Dr. Aykut BEKE for their endless support and undeniable trust during the semester. \textbf{\textit{smq}}. }           
\summary{

Otonom araçlar, yatırımlar yapıldıkça gerçeğe dönüşmektedir. Son yirmi yılda, otonom araçlar askeri, lojistik ve endüstrideki potansiyel kullanımları nedeniyle akademisyenler ve iş dünyasından büyük ilgi gördü. Araçların otonomluğunun farklı yönlerden faydaları vardır. Trafikte meydana gelen kazaların çoğu insan hatası olduğundan, otonom araçlar insan hatalarından kaynaklanan kazaların sayısını önemli ölçüde azaltabilir. Ayrıca araç sürücüsünün yolculuk sırasında farklı şeylerle uğraşmasına imkan sağlayacaktır. Tabii ki araçların bu otonomluğu kısa bir süre içerisinde doğrudan kullanılmayacak ancak sabit seyir hızı veya şerit takip sistemleri gibi otonom araç özellikleri halihazırda yaygın olarak kullanılıyor ve araçların otonom özellikleri birer birer hayatımıza giriyor. Trafik gibi karmaşık ve dinamik bir ortamda tamamen otonom olarak hareket etmek çözülmesi zor bir problemdir. Bu proje kapsamında otonom sürüşün bir diğer önemli özelliği olan bu zor probleme mini bir araba üzerinde çözüm geliştirilmeye çalışılacaktır. BU projede gerçek dünyanın bir simülasyonu olarak küçük bir araç için bu sorunun üstesinden gelmek amaçlanmaktadır. Robotik araştırma ve öğretim için karmaşık bir açık kaynaklı platform olan MIT RACECAR(Rapid Autonomous Complex-Environment Competing Ackermann-steering Robot), son teknoloji sensörler ve işleme yeteneklerine sahip olan bir araçtır. Son birkaç yıldır, güvenilirlik ve maliyet gibi alandaki ilerlemeler ve araştırmalar üzerindeki çeşitli kısıtlamalara rağmen, uygun fiyatlı mobil platformlar ve 1/10 ölçekli robotik platformlar piyasaya sürüldü. Ancak, MIT RACECAR, tamamen açık kaynaklı ve ROS(Robot Operating System) kullanan standartlaştırılmış sistemler kullanılarak oluşturulan Ackermann direksiyonuyla gerçekçi araç dinamikleri sunarak araştırma ve öğretim için sağlam bir robotik platform olarak öne çıkmaktadır. Bu tezin amacı, MIT RACECAR platformu için çevresel değişkenlere dayanıklı ve gerçek zamanlı olarak başarılı bir şekilde çalışabilen bir rota planlama algoritması oluşturmak ve ayrıca engellerin olduğu karmaşık ortamlarda güvenli araç seyahati sunmaktır. Platform, ZED kamera, NVDIA Jetson TX2 geliştirici kartı, lidar gibi donanımların yanı sıra, Ubuntu ve ROS'un birlikte çalışabilir bir ortamını da sağlayarak tez konusu olan rota planlaması için uygun bir platform sunmaktadır. Aracın hareketini yöneten kinematik ilişkileri analiz etmek için, MIT RACECAR'da rota planlama uygulanmadan önce bir araç kinematik modeli geliştirilmiştir. Bu amaçla geçici hal cevabını iyileştirmek için, düşük seviyeli kontrolör yapısı modellendi ve MIT RACECAR'ın performansı, gerçek zamanlı deneylerle incelendi.

Projede ilk olarak aracın ROS ile kontrolü sağlandı. BU amaçla joystick ile kontrol edilmesi için gerekli düğümler hazırlandı. SOnrasında sırasıyla DWA(Dynamic Window Approach), TEB(Timed-Elastic Band) ve APF(Artificial Potential Field) rota planlama algoritmaları MIT RACECAR'a uygulandı. Bu algoritmaların biribirlerine karşı farklı konularda avantajları ve dezavantajları bulunmaktadır. Bu sebeple algoritmaları karşılaştırmak için bir senaryo oluşturuldu. Bu senaryoya göre oluşturulan kıvrımlı çift şeritli bir yol üzerinde MIT RACECAR'ın şeritleri takip etmesi ve bir engelle karşılaştığında yoldan çıkmadan şerit değiştirerek engele çarpmadan geçmesi gerekmektedir. Bu senaryoyu uygulayabilmek için ihtiyaç duyulan şeritlerin konumu bilgilerini elde etmek için görüntü işleme tabanlı bir algoritma geliştirildi. Bu algoritma ZED camera dan aldığı görüntüyü işleyerek hedef noktayı tespit edip, hedef nokta bilgisini rota planlama algoritmasına vermektedir.

Gerekli araçlar oluşturulduktan sonra, algoritmalar senaryoya göre test edildi. Bu testlerde algoritmanın kaç engeli başarıyla geçtiği, ne kadar basit rotalar seçtiği, işlem yükü gibi ölçümler yapıldı.Bu sonuçlara göre en fazla engeli başarıyla geçen algoritma olmamasına rağmen işlem yükünün az olması ve basit çalışma mantığından dolayı APF seçildi. Kompleks olamayan yapısıyla APF nin projenin ileriki aşamalarında da avantaj sağlayacağına inanılmaktadır.

}         
\ozet{

Autonomous vehicles are really becoming a reality as investments are made. In the last two decades, autonomous vehicles have gotten a lot of interest from academics and business because of its potential uses in military, logistics, and industry. Autonomy of the vehicles has benefits from different aspects. It will allow the driver of the vehicle to deal with different things during the journey or for the safety concerns, since the most of the accidents happen in traffic are human mistakes, autonomous vehicles can significantly reduce the number of accidents caused by human mistakes. Of course, this autonomy of vehicles will not be used directly in a short time period, but autonomous vehicle features like speed cruise speed or lane tracking systems are already in use. However, autonomy of the vehicles are come into use one-by-one. Within this project, another important feature of autonomous driving will be tried to implement on a mini car. Moving completely autonomously in a complex and dynamic environment, like traffic, is a hard problem to solve. Although, this project aims to overcome this problem for a small vehicle as a simulation of the real world. On top of a small size race vehicle, the MIT RACECAR(Rapid Autonomous Complex-Environment Competing Ackermann-steering Robot), a complex open-source platform for robotics research and teaching, includes cutting-edge sensors and processing capabilities. Commercially available mobile platforms and 1/10 scale robotic platforms have been launched, despite various constraints on advancements and research in the field, such as reliability and cost. MIT RACECAR provides a robust robotic platform for research and teaching by offering realistic vehicle dynamics with Ackermann steering that is built using fully open-source and standardized systems that use ROS(Robot Operating System). The goal of this thesis is to create a path planning algorithm for the MIT RACECAR platform that is robust to environmental variables and can function successfully in real time, as well as to offer safe vehicle travel in complex environments with obstacles.The interoperability of the Ubuntu system with ROS offers a viable platform for path planning, which is the topic of the thesis, in addition to hardware such as the ZED camera, NVDIA Jetson TX2 developer board, and lidar on the vehicle. In order to analyze kinematic relations that govern the motion of the vehicle, a vehicle kinematic model is developed before route panning is implemented in MIT RACECAR. To avoid tracking accuracy concerns, the low-level controller structure was modeled, and the performance of the MIT RACECAR was examined with on-the-ground experiments.

In the project, the vehicle was first controlled with ROS. For this purpose, the necessary nodes were prepared to be controlled with a joystick. Afterwards, DWA(Dynamic Window Approach), TEB(Timed-Elastic Band) and APF(Artificial Potential Field) path planning algorithms were applied to MIT RACECAR, respectively. These algorithms have advantages and disadvantages against each other on different issues. For this reason, a scenario was created to compare algorithms. On a curved double lane road created according to this scenario, MIT RACECAR has to follow the lanes and when it encounters an obstacle, it has to change lanes without leaving the road and pass without hitting the obstacle. In addition, an image processing algorithm was developed to obtain the position information of the lanes needed to implement this scenario. This algorithm detects the target point by processing the image taken from the ZED camera and gives the target point information to the path planning algorithm.

After the necessary tools were created, the algorithms were tested against the scenario. In these tests, measurements such as how many obstacles the algorithm successfully passed, how simple routes it chose, and computational costs they have. According to these results, although it was not the algorithm that successfully passed the most obstacles, APF was chosen due to its low processing load and simple working logic. It was believed that with its uncomplicated structure, APF would also provide advantages in the future stages of the project. 
}

\begin{document}

\chapter{INTRODUCTION}

Autonomous vehicles are a rapidly expanding market and research field. In recent years, the automotive industry has increased its projects and importance in this area in order to reduce the number of deaths caused by crashes \cite{1_importance}. Furthermore, according to  \cite{1_economic}.the autonomous vehicle market is expected to grow 36.9\% between 2017 and 2027 and reach \$126.8 billion dollar market value. And the fact that it is such a big economy enables it to renew itself as something. In this direction, studies are carried out for smartening and autonomy in the design of automobiles. According to a World Health Organization research, road traffic collisions are the seventh leading cause of mortality across all age categories, accounting for about 1.35 million deaths in 2016, with cyclists, motorcyclists, and pedestrians contributing for more than half of the deaths \cite{1_who}. Considering the complex traffic situations and the increase in traffic accidents due to carelessness, misbehavior, fatigue or distraction, the improvements seen in different levels of autonomous vehicles are motivated to reduce injuries and deaths caused by complex traffic scenarios and errors. Injuries and deaths caused by traffic accidents can be reduced with more accurate Advanced Driver Assistance System(ADAS) applications.

In light of all of this knowledge, in order to be able to prevent crashes or pass through an obstacle, path planning algorithms are being developed for decades. Path planning algorithms are basically used to enable autonomous vehicles to move in environments with obstacles. Also, as mobile robots have recently started to work in dynamic environments where people are also present, path planning algorithms have gained more importance in order for autonomous vehicles to work safely. To achieve autonomous driving skills, different path planning algorithms use different approaches. While algorithms like RRT(Rapidly-Exploring Random Tree), RRT*(Rapidly-exploring random tree (star)) and PRM(Probabilistic Roadmap) use probabilistic methods for path planning, algorithms like APF use geometric approaches.

\clearpage

When it comes to real-world applications, such as autonomous vehicle deployment, testing the boundaries of safety and performance on real cars is expensive, inaccessible, and dangerous. Commercially available mobile systems such as Jackal UGV(Unmanned Ground Vehicle) \cite{1_jackal} and TurtleBot2 \cite{1_turtle} have been developed to resolve these concerns. Several small-scale robotic platforms have been built in recent years to further research, particularly in the area of autonomous driving. Many of these platforms are built on a small-scale racecar with a mechanical framework to support the electronic components, generally 1/10 size of an actual vehicle. In 2014, Mike Boulet, Owen Guldner, Michael Lin, and Sertac Karaman developed RACECAR (Rapid Autonomous Complex-Environment Competing Ackermann steering Robot) that was the first mobile robot with a strong graphics processing unit(GPU)  \cite{1_sertac}. By offering realistic car dynamics with Ackermann steering and drive trains, the RACECAR platform provides as a robust robotic platform for research and education. It is built on completely open-source and standardized systems that use ROS and its related libraries.

\section{Purpose of The Thesis}

This project mainly focuses on implementing and testing different path planning algorithms on MIT RACECAR platform. These algorithms are expected to avoid obstacles while following the road on a high curvature road. It is expected to determine the path planning algorithm that implements this scenario most successfully and to create a starting point for the solution of complex problems such as overtaking and escape maneuver using this algorithm in the period after the thesis. A lane finding algorithm based on image processing had to be developed at the same time in order to implement the scenario. Within this project, necessary tools will be developed and, route planning algorithms will be evaluated by creating a scenario environment.

\clearpage

\section{Literature Review}

Since path planning is a subject that has been studied for a long time, many algorithms related to path planning have been developed. DWA is one of the most popular algorithms. Fox proposed DWA algorithm in 1997 but, it is still frequently used. Also, algorithms that relies on geometric relations like APF are very successful with a low computational cost \cite{1_apf}. Also, algorithms based on optimization of dynamic parameters like TEB are proposed by Rösmann and it is shown that it can navigate complex environments \cite{3_teb_2}. Several vision-based lane detection algorithms have been presented in the literature to reliably detect lanes. These approaches may be divided into two types: feature-based and model-based. Detecting the evident properties of lanes that distinguishes them from the rest of the objects on the lane image is the basis of the feature-based technique. The task of feature detection would be split into two parts: edge detection and line detection. To identify lane edges and model lane borders, edge-based techniques have been developed. The accuracy and computing needs of more sophisticated implementations of these models rise, while the resilience to noise decreases. For feature extraction, the Canny edge detection method is presented, which offers an exact match to lane lines and is adaptable to a complex road environment \cite{1_canny}. Improvements to Canny edge detection can successfully deal with numerous noises in the road environment, according to \cite{1_canny2}. Furthermore, defining the targeted region of the picture where lane borders are located, called to as region of interest (ROI), enhances performance efficiency \cite{1_roi}. To ease the identification procedure, the region of interest would be split into left and right sub-regions based on the width of the lane.

\chapter{Platform Overview}

\begin{figure}[h!]   
  \centerline{\includegraphics[width=.6\linewidth]{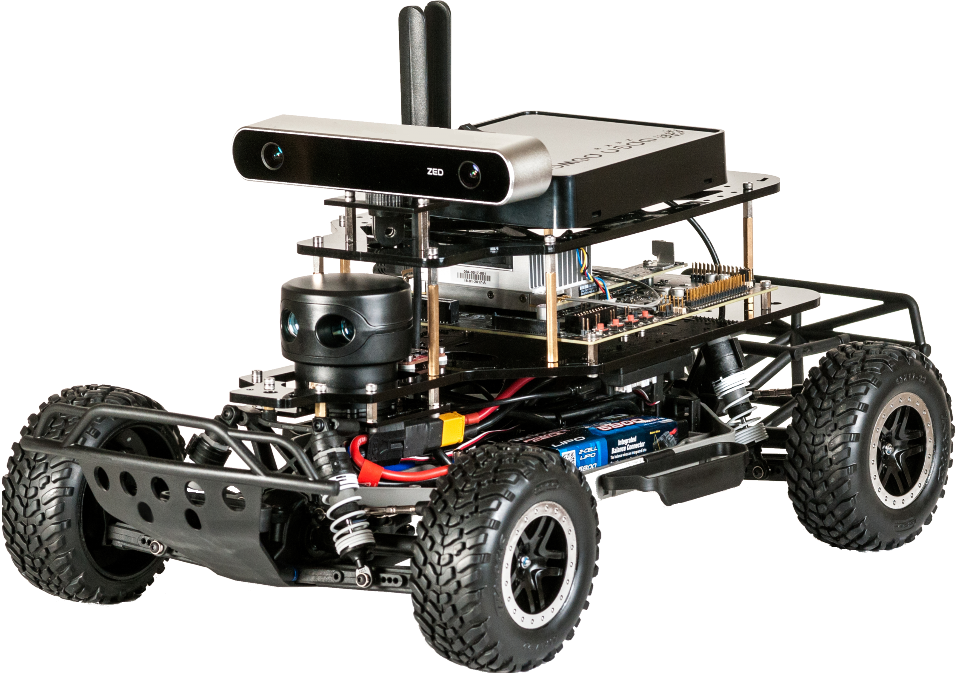}}
  \vspace{5mm}
  \caption{MIT Racecar Platform}
  \label{fig:racecar}
\end{figure}

RACECAR is a full stack robotics system that provides high computing power with reliable mechanical structure. RACECAR platform serves as a strong robotic platform for research and education by providing realistic vehicle dynamics with Ackermann steering, and drive trains. It is designed based on fully open-source and standardized systems that take advantage of ROS and its associated libraries \cite{2_racecar}. MIT RACECAR has sensors that can be listed as:

\begin{itemize}
    \item StereoLabs ZED Stereo Camera
    \item RPLidar A2 Lidar
    \item Sparkfun 9DoF Razor IMU
\end{itemize}

Because of its numerous sensors and well documented software, MIT RACECAR was chosen to be used in this project as the robot that trajectory generation algorithms will be tested on.

\clearpage

\section{Hardware}

\subsection{Chasis}

Chassis of MIT Racecar platform is Slash 4×4 Platinum Truck model from Traxxas. These chassis is giving providing enough performance parameters and has enough space for installation of other equipments. The chassis is able to drive 40 mph with its default Velineon 3500 Brushless Motor, but VESC software the maximum speed is limited by 5 mph for safety concerns. Power required by BLDC is provided by a 2S 4200 mah LiPo battery \cite{2_traxxas}.

\subsection{NVIDIA Jetson TX2}
NVIDIA Jetson TX2 is high power graphical processing unit which is widely used in robotics applications. It has enough processing power for trajectory planning and other tasks. Jetson TX2 has an ARMCortexA57 CPU(Central Process Unit) and 256 CUDA enabled GPU with an 8 GB RAM(Random Access Memory) \cite{2_jetson}.

\begin{figure}[h]   
  \centerline{\includegraphics[width=.5\linewidth]{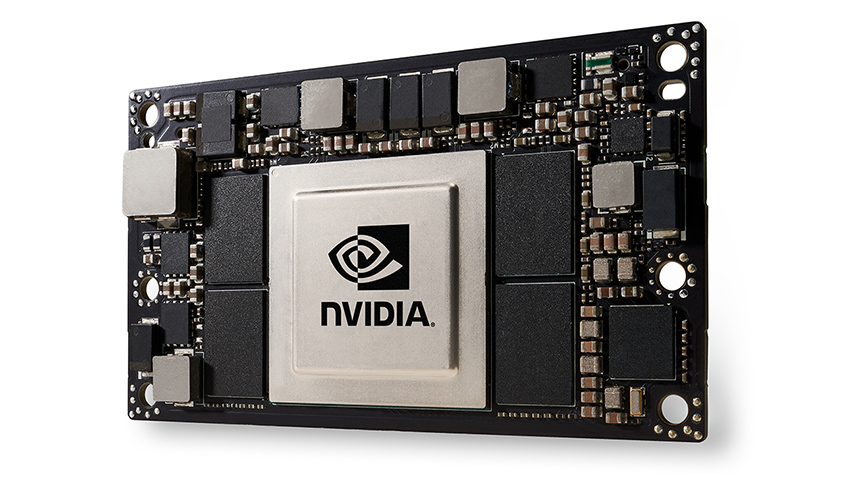}}
  \caption{NVIDIA Jetson TX2 \cite{2_jetson}}
  \label{fig:racecar}
\end{figure}

\subsection{StereoLabs ZED Stereo Camera}
StereoLabs ZED Stereo Camera is one of the enviromental awareness sensors MIT Racecar. ZED camera comes with dual 4 MP Camera which that provides 110° FOV(Field of View). ZED camera provides real-time pointcloud data in addition to getting 1080p HD video at 30 FPS or WVGA at 100 FPS(Frame Per Second) thanks to its SDK(Software Development Kit) \cite{2_zed}.

\clearpage

\begin{figure}[h!]   
  \centerline{\includegraphics[height=3cm]{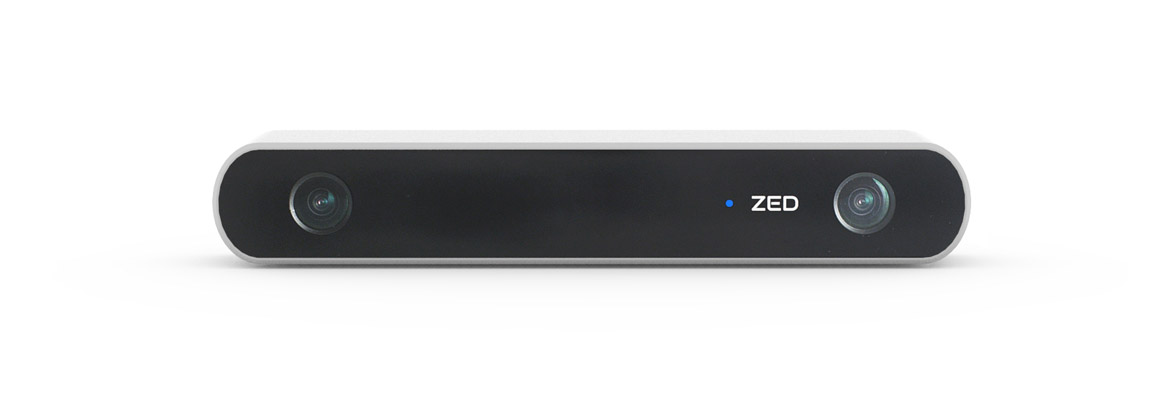}}
  \caption{StereoLabs ZED Stereo Camera \cite{2_zed}}
  \label{fig:racecar}
\end{figure}

\subsection{RPLidar A2 Lidar}

RPLidar A2 Lidar is a low-cost 2d lidar that developed by SLAMTEC will give us the obstacle distance information which is necessary for trajecrory generation. It can measure distances with a 360 degree field of view with maximum 16 meters range \cite{2_lidar}.

\begin{figure}[h]   
  \centerline{\includegraphics[width=.7\linewidth]{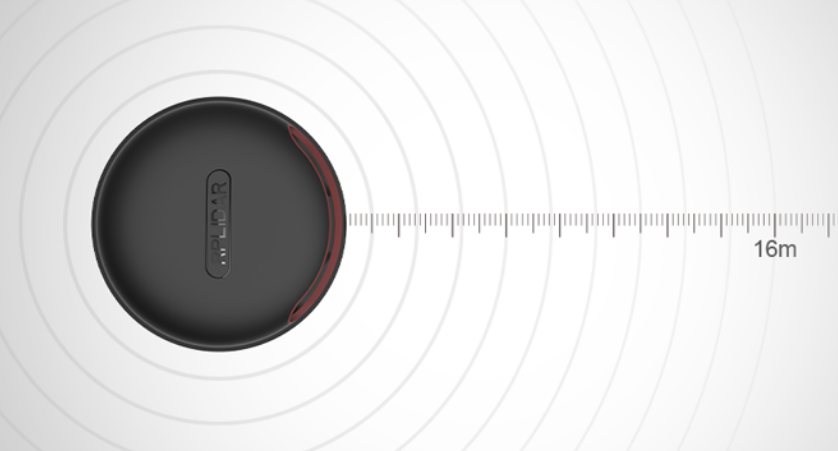}}
  \vspace{5mm}
  \caption{RPLidar A2 Lidar \cite{2_lidar}}
  \label{fig:racecar}
\end{figure}

\subsection{Sparkfun 9DoF Razor IMU}

Sparkfun 9DoF Razor IMU is reprogrammable IMU(Inertial Measurement Unit) with an MPU-9250 9DoF (9 Degrees of Freedom) sensor. Sparkfun IMU is a easy-to-use IMU with its various connection types like UART, I2C and USB. In MIT RACECAR platform it is fully entegrated by a USB connection but in this project it is not actively used \cite{2_imu}.

\subsection{VESC Electronic Speed Controller}

BLDC(Brushless DC Motor) and steering servo of the MIT Racecar is controlled by Vedder Electronic Speed Controller (VESC). Unlike conventional ESCs, VESC also provides useful data over USB connection like current and voltage data for each motor phase and wheel odometer data. VESC is capable of controlling speed and steering angle of MIT Racecar succeffully. These low level controllers of VESC allow researchers to be more focused on developing solutions to higher level problems.

\section{Software}

The MIT RACECAR software relies on JetPack and ROS as two main components. JetPack is a special software for NVIDIA Jetson computers that specializes Ubuntu for them. Along with specialized OS for Jetson it also provides libraries, APIs, samples and developer tools for AI applications and GPU computing \cite{2_jetpack}.

As the other main component of The MIT RACECAR software, RACECAR ROS package provides robotic capabilities on ROS as subpackages and it consists of sub-packages like; \textit{\textbf{ackermann\_cmd\_mux, joy\_node rplidar\_ros, sparkfun-9dof-razor-imu, zed\_ros\_wrapper.}} These packages are required for collecting sensor data from vehicle, applying commanded control signals,  making necessary transformations between different frames. As an addition to these packages, for data collection and logging \textit{\textbf{rosbag}} package is used. The rosbag package allows us to save the sensor data as time-series and then play it back in real time. \textit{\textbf{Rosbag}} package also allows us to measure performance of developed algorithm. Also, for visualization purposes, another ROS package \textit{\textbf{rqt}} is used.

\subsection{Robot Operating System (ROS)}

ROS is an open-source framework that mainly handles communication between nodes and creates a baseline for researchers. Since it also has useful packages like transformation calculations etc., it is very helpful for beginners in robotics programming and accelerate prototyping processes for researchers.

\begin{figure}[h!]   
  \centerline{\includegraphics[width=.7\linewidth]{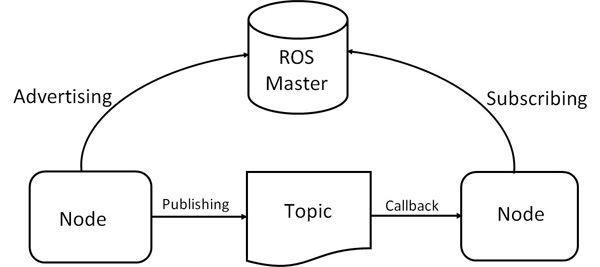}}
  \vspace{5mm}
  \caption{ROS Topics Communication System}
  \label{fig:ros_communication}
\end{figure}

\clearpage

Communication architecture of ROS is communication between nodes with a publish-subscribe messaging model. This communication is only possible with a \textbf{\textit{master}} which serves as an XML-RPC based server. \textbf{\textit{Master}} makes the communication between nodes with an address which is named \textbf{\textit{topic}}. \textbf{\textit{Topics}} are special messaging addresses with a specific data type and name. 

The ROS nodes of MIT RACECAR can be seen in Table \ref{ros_node_t}.

\begin{table}[]
\centering
\begin{tabular}{@{}
>{\columncolor[HTML]{EFEFEF}}c 
>{\columncolor[HTML]{EFEFEF}}c 
>{\columncolor[HTML]{EFEFEF}}c @{}}
\toprule
\textbf{NODE NAME} &
  \textbf{PACKAGE} &
  \textbf{DESCRIPTION} \\ \midrule
\begin{tabular}[c]{@{}c@{}}ackermann\_cmd\\ \_mux\end{tabular} &
  \begin{tabular}[c]{@{}c@{}}ackermann\_cmd\\ \_mux\end{tabular} &
  \begin{tabular}[c]{@{}c@{}}Ackermann velocity command inputs are \\ multiplexed in this node. It controls \\ incoming ackermann command topics but,\\  based on priority, allows \\ one topic command at  a time.\end{tabular} \\ \midrule
joy\_node &
  joy &
  \begin{tabular}[c]{@{}c@{}}This node is used for generic joystick to ROS. \\ It publishes joy message that \\ includes joystick all possible states \\ of buttons and axes.\end{tabular} \\ \midrule
rplidarNode &
  rplidar\_ros &
  \begin{tabular}[c]{@{}c@{}}Driver for RPLIDAR sensor that converts messages \\ that comes from lidar sensor to laser scan messages \\ and publishes to a topic.\end{tabular} \\ \midrule
\begin{tabular}[c]{@{}c@{}}zed\_camera\\ \_node\end{tabular} &
  zed-ros-wrapper &
  \begin{tabular}[c]{@{}c@{}}Driver for ZED camera that outputs pointcloud, \\ camera left and right images and pose \\ information of robot.\end{tabular} \\ \midrule
rosbag\_node &
  rosbag &
  \begin{tabular}[c]{@{}c@{}}It is a tool for recording the messages published on \\ topics to bag files which are specially formatted files \\ that stores timestamped ROS messages.\\ It also used for replaying the recorded topics.\end{tabular} \\ \midrule
rqt\_node &
  rqt &
  \begin{tabular}[c]{@{}c@{}}This node is used for visualizatioon purpose. \\ It provides tools to easily plot data in real-time,\\ displays rostopic messages etc.\end{tabular} \\ \bottomrule
\end{tabular}
\caption{The ROS nodes of MIT RACECAR software package}
\label{ros_node_t}
\end{table}

\clearpage

\section{Vehicle Kinematic Model}

Ackermann steered vehicles like MIT RACECAR are commonly expressed as bicycle kinematic model. Bicycle kinematic model for Ackermann steered vehicles assumes that front wheels and rear wheels are combined amongst themselves and end up with a two-wheel bicycle \cite{2_bicycle}. In addition to this, bicycle kinematic model also assumes that the vehicle only moves on XY plane and thanks to these assumptions, bicycle kinematic model is a simple geometric relationship between the steering angle and the curvature that the rear axle of the vehicle will follow.

\begin{figure}[h]   
  \centerline{\includegraphics[height=.6\linewidth]{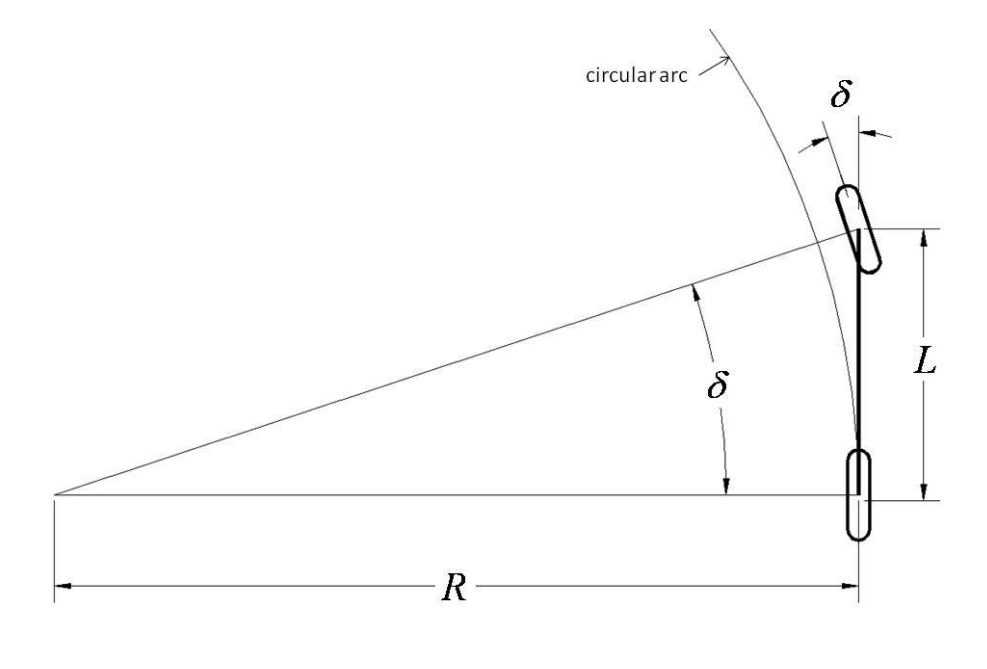}}
      \caption{Simple Bicycle Model \cite{2_bicycle}}
  \label{fig:bicycle}
\end{figure}

\begin{equation}
    tan(\gamma) = \frac{L}{R}
\label{eq_bicycle}
\end{equation}

The geometric relationship can be written as equation \ref{eq_bicycle} where $\gamma$ is steering angle, $L$ is the wheelbase and $R$ is the turning radius of the vehicle. It should be noted that this geometric relationship gives acceptable results for only low speed and moderate steering angles.

\clearpage

\chapter{PATH PLANNING}

Path planning algorithms are used to enable mobile robots to perform their most basic tasks, which are their ability to move. These algorithms mainly consists of two parts which is path planning and trajectory planning. While path planning is responsible for the higher level motion of the robot, trajectory planning algorithms calculates the required actuator behavior for following via-points generated by path planning \cite{3_global_local}. This project focuses on trajectory planning algorithms. In ROS methodology, path planning algorithms corresponds to global planner and trajectory planner corresponds to local planner.

\begin{figure}[h!]   
  \centering
  \centerline{\includegraphics[width=\linewidth, height=14cm]{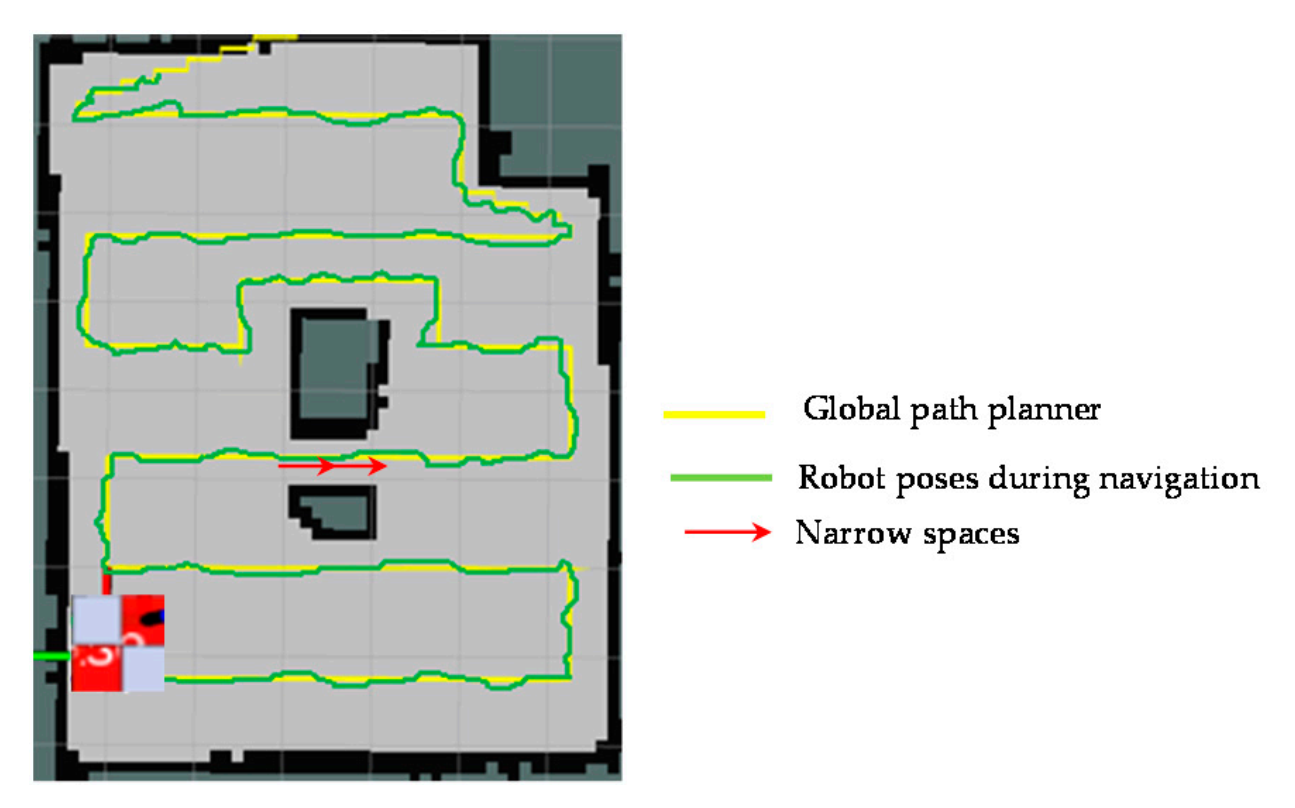}}
  \caption{An example view of local and global planners \cite{3_global_local_fig}.}
  \label{fig:costmap}
\end{figure}

\clearpage

In this project, the performance of different trajectory planning algorithms like DWA, TEB and APF, will be compared by their performance. In the realization phase of the project, the algorithms will be tested on the Racecar Platform in a curvy road. The algorithms are implemented in the ROS environment. Thanks to the useful features of ROS, implementation and testing of the algorithms could be done easily.

\section{Main Elements of Path Planning}

Path planning processes may become complex processes. In order to make path planning more tidy and less complex, some auxiliary ROS packages will be used. Before diving into path planning algorithms, these auxiliary packages will be examined.

\subsection{Global Planner}

Global planners are responsible for creating a collision free path for the robot. Global planners generate this path around the obstacles from a geometric approach, and they don't take into account  vehicles dynamics. Unlike local planners, generally they plan for a larger map. A*, Dijkstra's Algorithm, RRT, $RRT^*$, and PRM can be counted as examples of global planning algorithms. The output of Global planners is a roadmap or waypoints to the target point. These roadmap provides local planner the points to follow.

In this project, \textbf{\textit{base\_global\_planner}} package from ROS navigation stack will be used as global planner. This package provides a good baseline that can be configured to use different algorithms for global planning. In this project, since global planning is not one of the main tasks of this project, the package is configured for the Dijkstra's algorithm for calculating the shortest path that has the lowest cost without including robot kinematic parameters. Dijkstra's algorithm is mainly chosen for its simplicity and being computationally cheap. Global planners are basic algorithms for choosing the path that has minimum costs. Though, their performance heavily depends on configuration of global costmap. Since there is no mapping in this project, we are using our global planner with respect to the base\_link frame.

\clearpage

\subsection{Local Planner}

Local planner is the planner that calculates the short term plan and executes the plan. Unlike global planner, local planner includes robot kinematic parameters and calculates the plan according to these parameters. This extra calculation comes with a computational cost. In order to decrease the computational cost of local planners, boundaries of the local planner generally being selected small and local planner aims the furthest point that is on the global plan that is inside of local planner boundaries, not the real target point. 

Global planners produce waypoints or roadmaps as output and local planners get the output of global planners as inputs and calculates the required control signal to follow these waypoints. This control signal is generally a ROS message type named \textbf{\textit{"/geometry\_msgs/Twist"}}. The ROS message contains the commanded linear and angular velocity data, and another node takes this message and controls the vehicle with respect to these commanded velocities.

\begin{figure}[h!]   
  \centerline{\includegraphics[width=.8\linewidth]{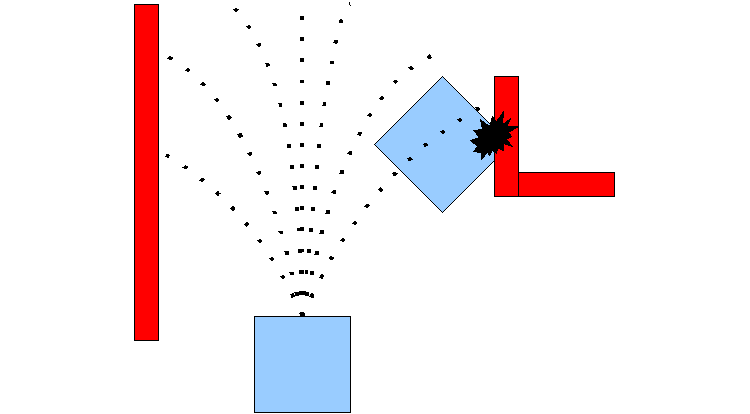}}
  \vspace{5mm}
  \caption{Local Planner trajectory generation \cite{3_local_planner}.}
  \vspace{5mm}
  \label{fig:planner}
\end{figure}


    
    
    
    

\subsection{Global and Local Costmaps}

Cosmaps are basically maps that stores the costs of every point on the map that is calculated based on sensor measurements or provided by a known map. 2D and 3D costmaps can be constructed, but for ground vehicles like the MIT RACECAR 2D costmaps are more suitable since the movement in z-axis can be neglected. 

\clearpage

\begin{figure}[h!]   
  \centering
  \includegraphics[width=\linewidth]{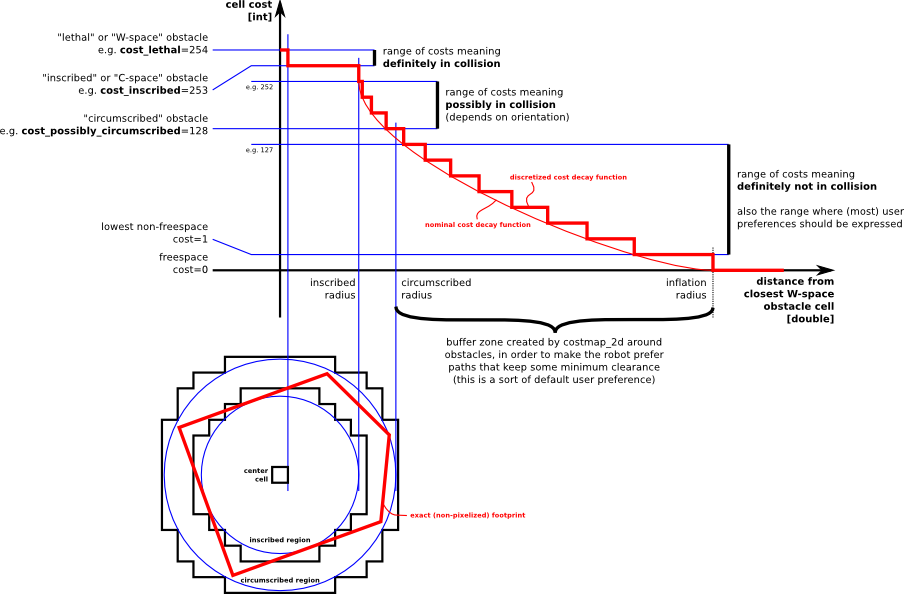}
  \vspace{5mm}
  \caption{The effect of inflation parameters on cell cost \cite{3_ros_costmap}}
  \vspace{5mm}
  \label{fig:inflation}
\end{figure}

Mainly, the costs of cells in costmaps are calculated based on whether there is an obstacle on that cell. In addition to this, another source of cost is the distance of the cell to an obstacle. This "distance" is named \textbf{\textit{inflation}} in ROS Navigation stack. The inflation is calculated by parameters called \textbf{\textit{inflation\_radius}} and \textbf{\textit{cost\_scaling\_factor}}. These parameters are required for deciding whether the cell will be determined as near an obstacle or not, and decay rate of inflation cost. The effect of inflation parameters on the cost of a cell can be seen in Figure \ref{fig:inflation}.

\begin{figure}[h!]   
  \centerline{\includegraphics[width=\linewidth]{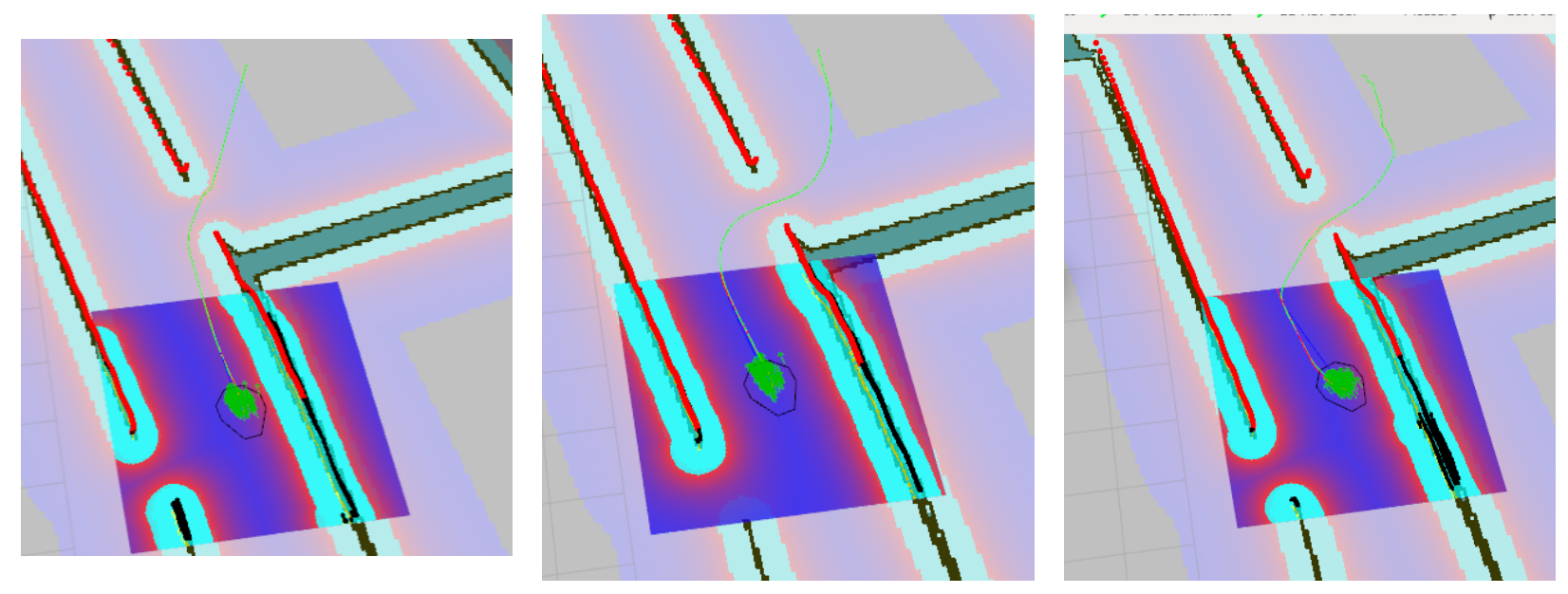}}
  \vspace{5mm}
  \caption{The effect of cost factor on generated path (cost\_factor=0.01, cost\_factor=0.55, cost\_factor=3.55) \cite{3_tuning_guide}.}
  \vspace{5mm}
  \label{fig:cf_effect}
\end{figure}

\begin{figure}[h!]   
  \centerline{\includegraphics[width=\linewidth]{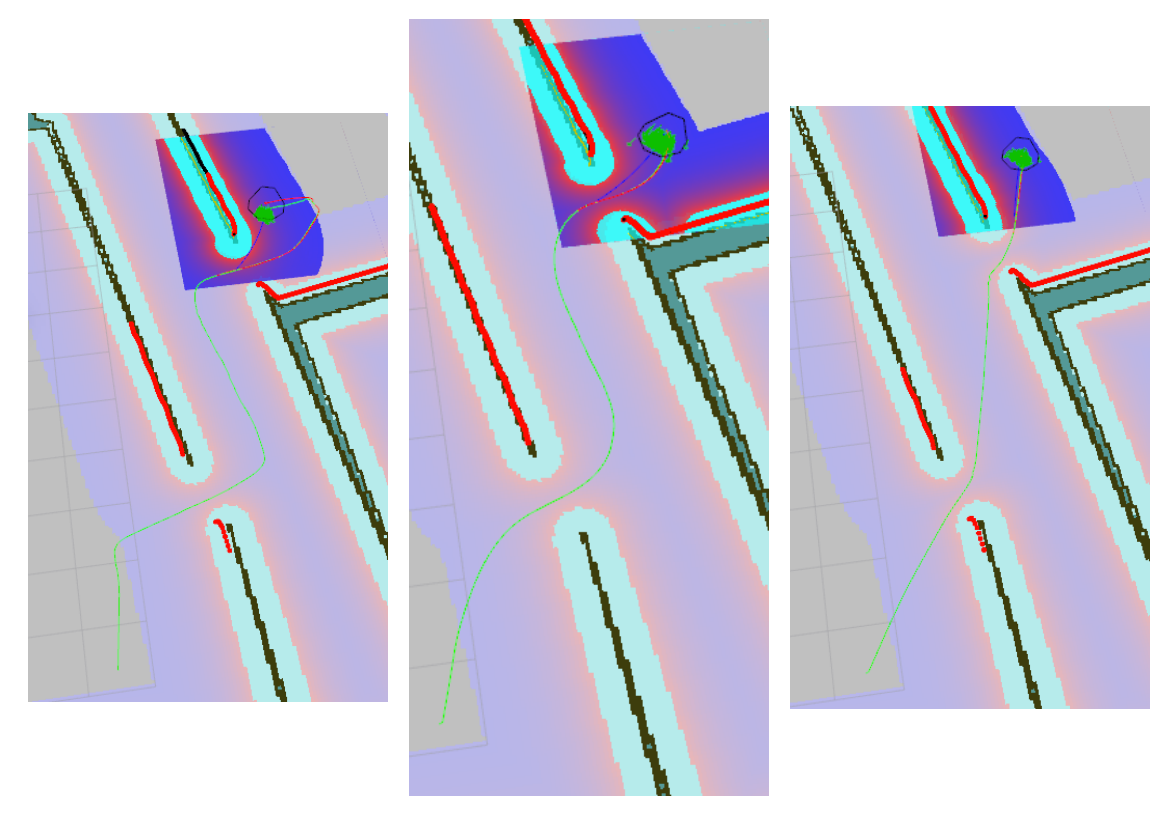}}
  \vspace{5mm}
  \caption{The effect of neutral factor on generated path (neutral\_factor=1, neutral\_factor=66, neutral\_factor=233) \cite{3_tuning_guide}.}
  \vspace{5mm}
  \label{fig:nc_effect}
\end{figure}

Also, another important parameters of costmaps are \textbf{\textit{cost\_factor}} and \textbf{\textit{neutral\_cost}}. These parameters determine the smoothness and the curvature of the calculated path. The effects of these parameters can be seen in Figure \ref{fig:cf_effect} and Figure \ref{fig:nc_effect}.

\clearpage

And lastly, it should be noted that, as the names suggests, global costmaps are for global planners and local costmaps are for local planners. Although, there is no difference between global costmaps and local costmaps in fact, their parameter configurations are differs. The map sizes and sensors are main elements that can be difference between them.

\clearpage

\section{Trajectory Planning Methods}

In this project, 3 different trajectory planning algorithms is considered and applied on MIT RACECAR platform. These algorithms are; Dynamic Window Approach(DWA), Time Elastic Band(TEB) and Artificial Potential Field(APF).

\subsection{Dynamic Window Approach}

DWA is a proven concept that is used for a long time in robotics. The method was first proposed in 1999 \cite{3_dwa}. And it is still a popular method for mobile robots. DWA provides a controller between global path and the robot. It calculates the cost function for different control inputs and searches for the maximum scored trajectory to follow. Thanks to its simplicity and easy implementation, DWA is a good choice as the first algorithm for trajectory planning. Basic working principle of DWA can be expressed as below.

\begin{enumerate}
    \item Take dx, dy and dtheta samples from control space 
    
    \item Predict next states from current states based on sampled dx, dy and dtheta.
    
    \item Score each trajectory from predictions with distance to obstacles, distance to the goal, distance to the global path, and speed and remove the trajectories that collide with obstacles.
    
    \item Choose the trajectory with the highest score and send the trajectory to the robot.
    
    \item Repeat until goal is reached.
\end{enumerate}

In order to be able to use DWA effectively, DWA needs to be configured properly. DWA has parameters that will configure robot configuration, goal tolerance, forward simulation, trajectory scoring and global plan. One of the important parameters of the DWA is sim time parameter. Sim time is essentially the time length that will DWA plan for, and this parameter heavily affects computation time of the trajectory. In tests, it is seen that when sim time is set to low values like 2.0 seconds or less, the performance of DWA is not sufficient for passing complex pass ways because it can not see what will happen after the sim time. This result in a suboptimal trajectory. Also, it should be noted that, since all trajectories generated by DWA is simple arcs, setting the sim time to high value like 5 seconds will result in long curves that are not very flexible. Thus, in order to achieve good performance with DWA, setting sim time to an optimum value is a must.

\begin{figure}[h!]   
  \centerline{\includegraphics[width=\linewidth]{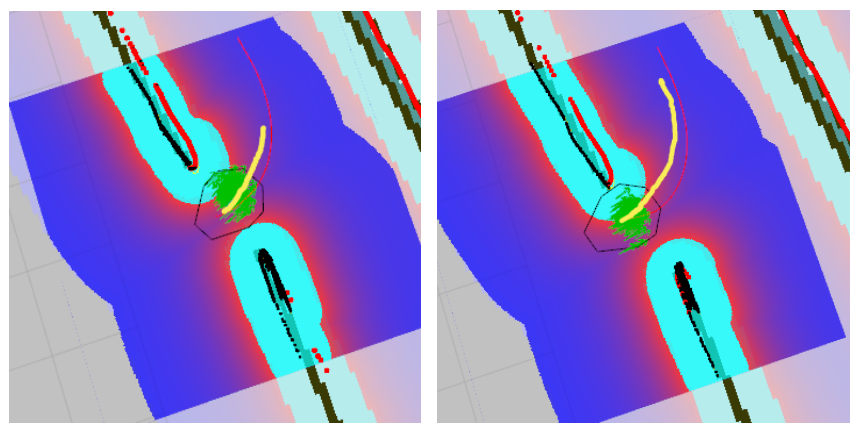}}
  \caption{The Effect of Simulation Time on Generated Trajectory (sim\_factor=1.5, sim\_time=4.0 \cite{3_tuning_guide}}
  \label{fig:planner}
\end{figure}

Aside from sim time, there are a few other factors to consider. Samples of velocity vx sample and vy sample are two parameters that determine how many translational velocity samples should be taken in the x and y directions, respectively. The number of rotational velocities samples is controlled by vth sample. The number of samples you want to take is determined by your computer's processing power. Because turning is generally more complicated than moving straight ahead, it is preferred to set vth samples to be higher than translational velocity samples in most cases. Since, our vehicle is non-holonomic there is no need for velocity samples in y direction. Simulation granularity is the step size between points on a trajectory that is referred to as sim granularity. It essentially means how often the points on this trajectory should be examined. A lower value indicates a higher frequency, which necessitates more processing power. 

Lastly, as an addition to all of these parameters, DWA can also plan both for holonomic and non-holonomic robots, but it does not support Ackermann drive robots as MIT RACECAR. Still, DWA was implemented on MIT RACECAR and got acceptable results.

\clearpage



\subsection{Time Elastic Band}

Time Elastic Band(TEB) planner was proposed by Rösmann as an improved version of elastic band algorithm \cite{3_teb}. The classic elastic band algorithm is based on optimizing a global path for the shortest path length. However, TEB optimizes the path by the time-optimal objective function. Also, while elastic band does take into account of dynamic constraints, TEB considers kino-dynamic constraints in the trajectory planning. Additionally, TEB planner also supports non-holonomic car-like robots.

TEB planner is essentially a solution to a sparse scalarized multi-objective optimization problem. The multi-objective problem includes constraints like maximum velocity, maximum acceleration, minimum turning radius etc. Since this optimization problem does not always have only one solution, TEB can get stuck in locally optimum points. Thus, sometimes the robot can not pass an obstacle even if there is a possible trajectory which the robot can follow. In order to solve this local minimum problem, an improved version of TEB was proposed \cite{3_teb_2}. With this improved version, TEB optimizes a globally optimal trajectory parallel to the multi-objective optimization problem that it already solves. The algorithm switches to this new globally optimum solution when necessary, and this way the local minimum problem of TEB is solved.

Due to being numerous kino-dynamic constraints for a car-like robot, TEB has numerous weights for each constraint and effects of weights on the generated trajectory depends on other weights. For this reason, optimizing TEB planner parameters requires good understanding of the concept, attention and controlled experiments. In this project, in  order to simplify this process, initial values of the TEB planner was optimized in simulation environment and final configuration fine-tuned in real time tests. The results showed that while MIT RACECAR can avoid obstacles smoothly with TEB planner, the planner still requires more tuning for getting better results while driving narrow areas.

\begin{figure}[h!]   
  \centerline{\includegraphics[width=\linewidth]{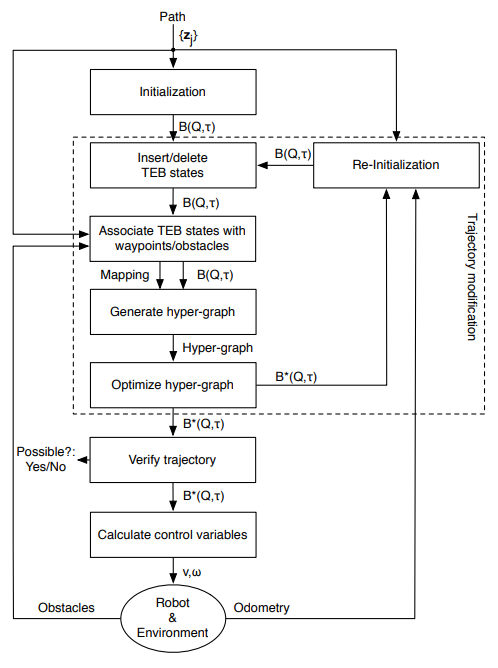}}
  \vspace{5mm}
  \caption{Flowchart of TEB Algorithm \cite{3_teb}}
  \vspace{5mm}
  \label{fig:planner}
\end{figure}

\clearpage

\section{Artificial Potential Field Method}

Artificial Potential Field(APF) Method is a basic approach for trajectory planning that is widely used by both industrial and academical applications. APF basically creates artificial attractive vectors that diverts vehicle to the goal position and repulsive vectors from obstacles that diverts vehicle from obstacles. Basically, addition of these attractive and repulsive vectors results in a target direction and speed for the vehicle.

\begin{equation}
    U(q) = U_{attractive}(q) + U_{repulsive}(q)
\end{equation}

And additionally, magnitudes of these vectors are defined by user. By tuning these magnitudes, the user can tune how close the robot will navigate through obstacles. With the help of this simple calculation, APF calculates how much it should divert from current heading.

As an addition to determining target orientation of the vehicle, speed of the vehicle can be calculated with the help of APF method by the density of obstacles. Simply, the density of obstacles determines the targeted velocity for the vehicle.

\begin{equation}
    V = V_{max} - K_{gain} \cdot n_{obstacles}
\end{equation}

\begin{figure}[h!]   
  \centerline{\includegraphics[width=.4\linewidth]{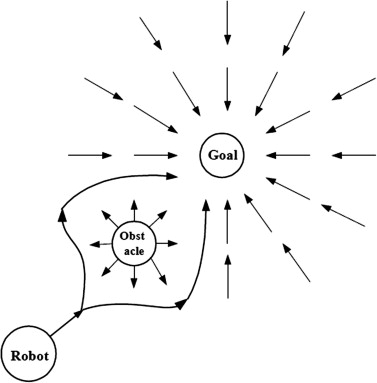}}
  \vspace{5mm}
  \caption{Basic trajectory generation of APF Method}
  \label{fig:planner}
\end{figure}

APF method has a very basic approach for trajectory planning. And this approach provides a non-complex solution to the problem with less computational costs. The main reason for choosing this method is this direct approach, but it has some problems that we will discuss in the next chapter.

\subsection{Problems of APF Method}

The main problem with the APF method is the local minimum problem. Similar to the previous version of TEB algorithm, APF can get stuck in local minimum points. As it can be seen in figure below, when repulsive forces from obstacles are equal in both directions, it calculates the resultant vector as the vehicle moves towards to the obstacle. This can result in a collision for the vehicle. Another problem of the APF method is Goal Non-reachable with Obstacles Nearby(GNRON) problem. In such cases that the goal point is near an obstacle, the repulsive force from the obstacle can prevent the robot from reaching the goal point by resulting in a local minimum that is close to the goal point.

These two problems are primary problems of APF method and various solutions are proposed like Evolutionary APF \cite{3_apf_evo}, Fuzzy APF \cite{3_apf_fuzzy} as modified APF algorithms.

\begin{figure}[h!]   
  \centerline{\includegraphics[width=.7\linewidth]{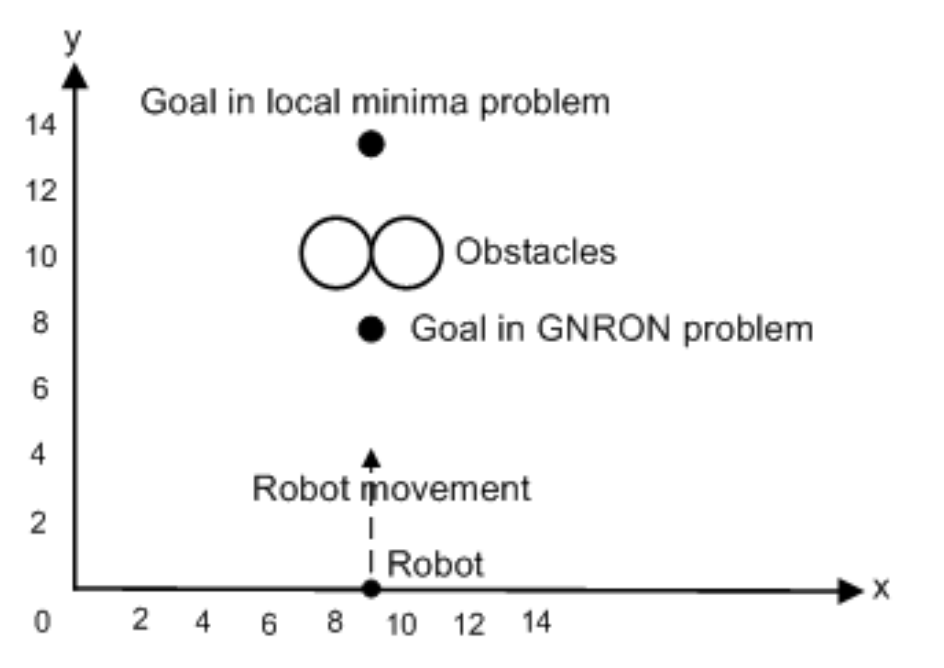}}
  \caption{Goal points with GNRON and local minimum problem \cite{3_apf_fig}}
  \label{fig:planner}
\end{figure}

While implementing APF to the MIT RACECAR, these problems were solved by implementing a local minimum detection method. This method, simply checks the calculated repulsive and attractive forces and if it determines that the robot is stuck in a local minimum point, it adds a vector that will divert the vehicle from going towards to the obstacle.

\chapter{Testing and Results}

Real life experimentation of trajectory planning algorithms on MIT RACECAR requires additional work. Until this part,  MIT RACECAR platform is inspected by its hardware and software. Also, numerous helper package of ROS is expressed and how trajectory planning algorithms can be applied. For the real time testing of different trajectory planning algorithms, there should be a measurement for determining how well the algorithm works. For that purpose, an environment was set up in Artificial Intelligence and Intelligent Systems(AI2S) laboratory. As the scenery, a double lane curvy road with obstacles in different locations was chosen. MIT RACECAR was expected to stay in lane and when it encounters to an obstacle, to pass the obstacle by changing the lane.

As a result of the chosen scenery, it is needed to detect lanes on the road. For achieving this, ZED Camera was used for image processing. Image processing in this project is achieved by OpenCV library. An image processing algorithm that detects lanes and returns goal points was designed with Python. Goal points are generated from detected lanes with respect to the look ahead distance that is determined by velocity of the vehicle dynamically.

Another aspect of the project is obstacle detection. Obstacle detection is done with the help of RPLidar A2 2D lidar. RPLidar provides users a ROS package that handles communication between computer and lidar and publishes obstacle information as a ROS topic. Lastly, all of these outputs are given to the path planning algorithm as target points and obstacles.

\section{Test Environment}

In order to create the required environment, a double lane curvy road was draw on the floor of AI2S Laboratory by electrical tapes. While making the road for MIT RACECAR, constraints like width of the vehicle, minimum turning radius of the vehicle are considered. Also, it is tried to avoid environmental effects like flare on the floor because of the lights. Since, the reference is being obtained from image processing, the flare is changing the quality of the reference signal in a bad manner and causes the vehicle to go out of the lane.

Also, the turning radius is an important constraint in such small environments, especially when obstacle avoidance is required. Thus, the curvature of the road was tried to be kept small enough to turn but also, big enough to see the limits of the algorithm.

Double lane structure of the road was chosen for lane changing when encountered by an obstacle. This structure also provides a good use case like lane changing and overtaking problems as future works. Besides, the middle lane of the road was chosen red in order to provide easier recognition of the lane. Since red color is easy to recognize with color filters like HSV, this provides a good starting point for the lane detection and leaves more time to focus on trajectory planning algorithms. The most painful part of the generated scenario for this study was that the changeable light conditions in the environment had a serious effect on the lane recognition algorithm's performance.

\begin{figure}[h!]   
  \centerline{\includegraphics[width=\linewidth]{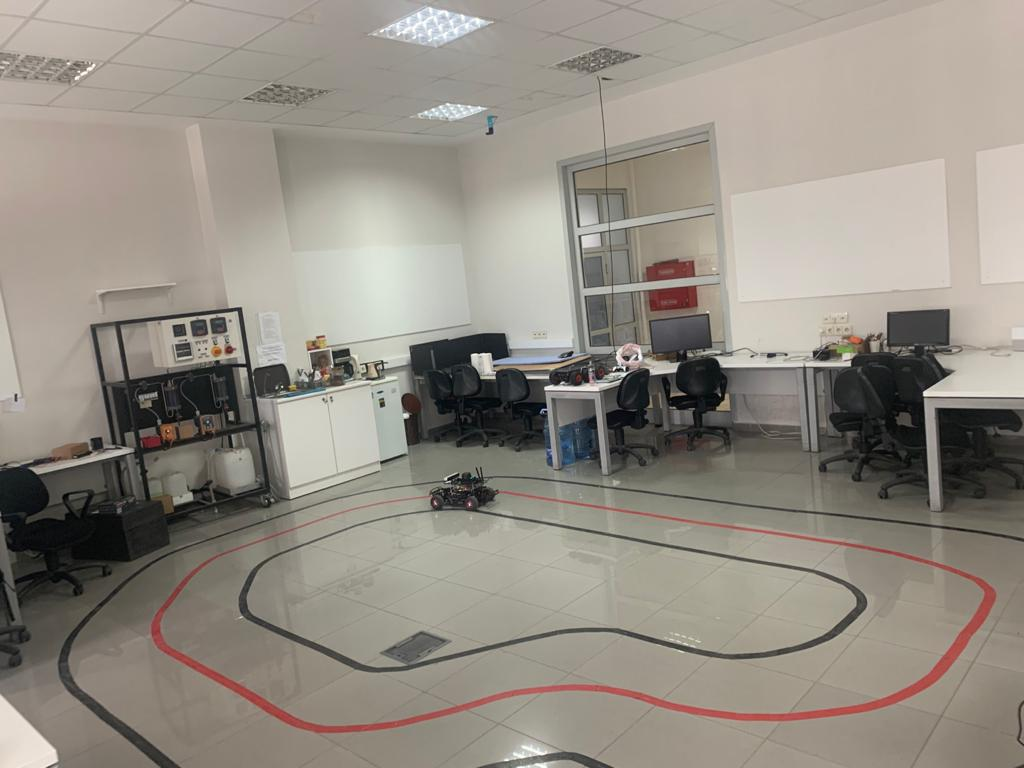}}
  \vspace{5mm}
  \caption{Testing environment}
  \label{fig:testing_env}
\end{figure}

\clearpage

\section{Lane Detection Algorithm}

Lane detection algorithm, which is a crucial part of the scenery that is chosen for the project, will be explained in this section. Lane detection algorithm consists of OpenCV functions that will not be explained since it is out of the scope of the project. The algorithm is designed to have two main parts for its being modular and easy to understand.

\begin{figure}[h!]   
  \vspace{5mm}
  \centerline{\includegraphics[width=\linewidth]{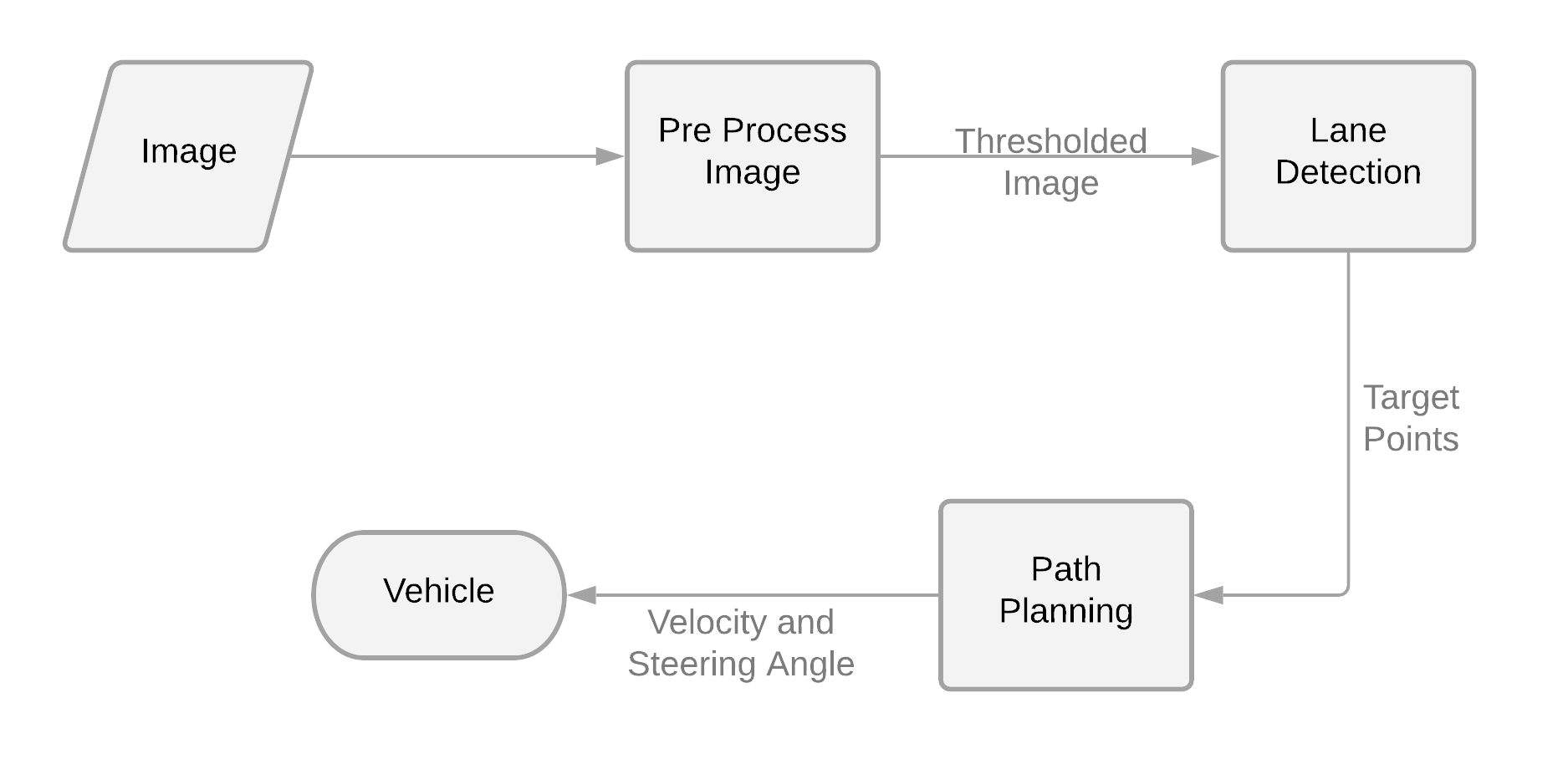}}
  \vspace{5mm}
  \caption{Main structure of the lane detection algorithm}
\end{figure}

\begin{figure}[h!]   
  \centerline{\includegraphics[width=\linewidth]{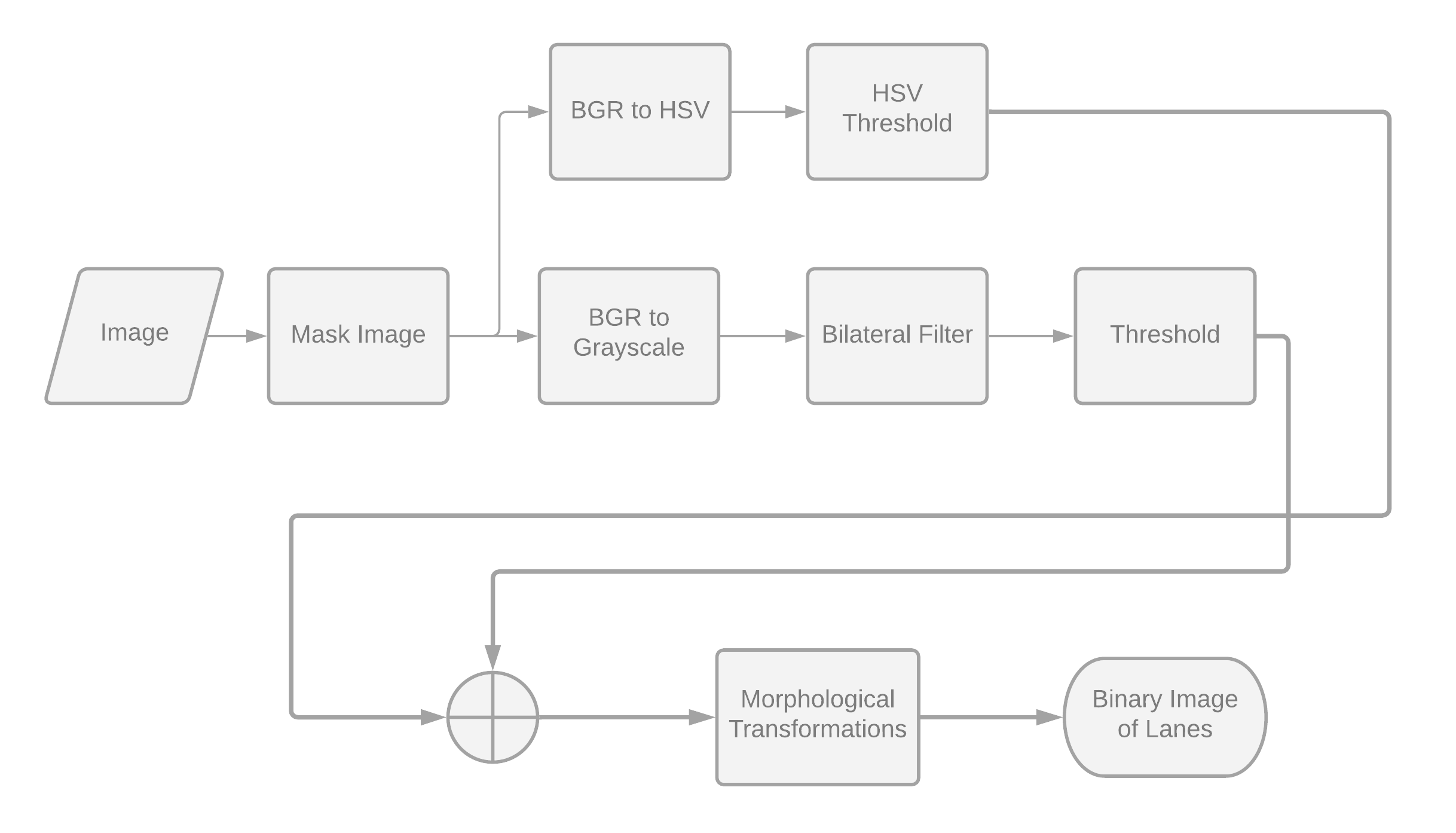}}
  \vspace{5mm}
  \caption{Flow of the first proposed image pre-processing algorithm}
  \label{fig:pre_1}
\end{figure}

The first part of the algorithm, the pre-processing part, is responsible for extracting lane information from camera input by eliminating everything except lanes itself from the image. The first idea for the algorithm was based on the idea to extract black lanes and red lane individually. For this purpose, a complex image processing algorithm that can be seen in Figure \ref{fig:pre_1} was proposed. However, the computational cost of this proposed method was not affordable for Jetson TX2. As a result of this problem, feedback loop of the system is updated only on 5-8 times average in one second and this situation was resulting in a hard control problem and slow response system.

\clearpage

In order to overcome these problems, a new, more plain algorithm that can be seen in Figure \ref{fig:pre_2} was proposed. The idea of the new algorithm is that it is not needed to find all the lanes individually, it is only needed to find red lane which is easier to find respectively. As a result, the new proposed method is less accurate but more effective and faster, and this loss of accuracy is a neglectable amount.

\begin{figure}[h!]   
  \centerline{\includegraphics[width=\linewidth]{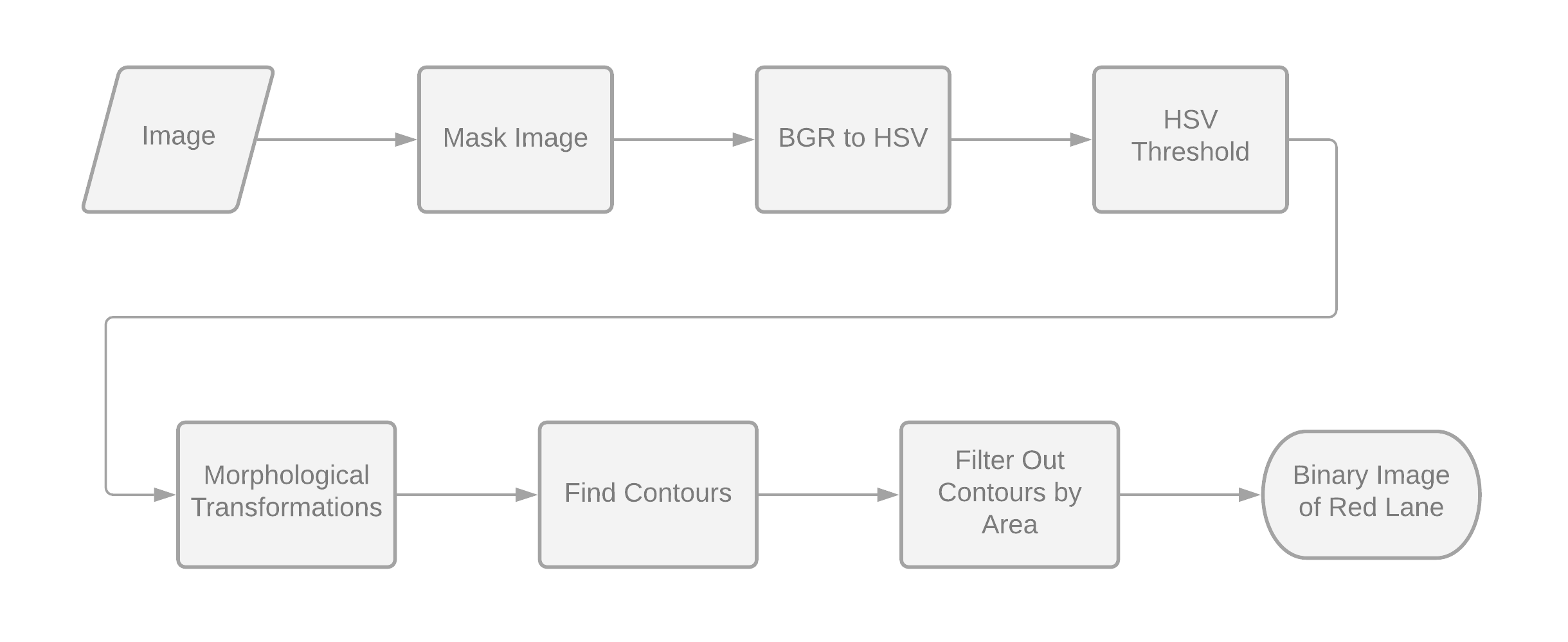}}
  \vspace{5mm}
  \caption{Flow of the final image pre-processing algorithm}
  \label{fig:pre_2}
\end{figure}

\clearpage

The pre-processing part of the algorithm uses mainly HSV masking, morphological operations and filtering contours based on sizes and area. The flow of the algorithm with an example input image can be inspected in Figure \ref{fig:pre_2}.

\begin{figure}[h!]   
  \centerline{\includegraphics[width=\linewidth]{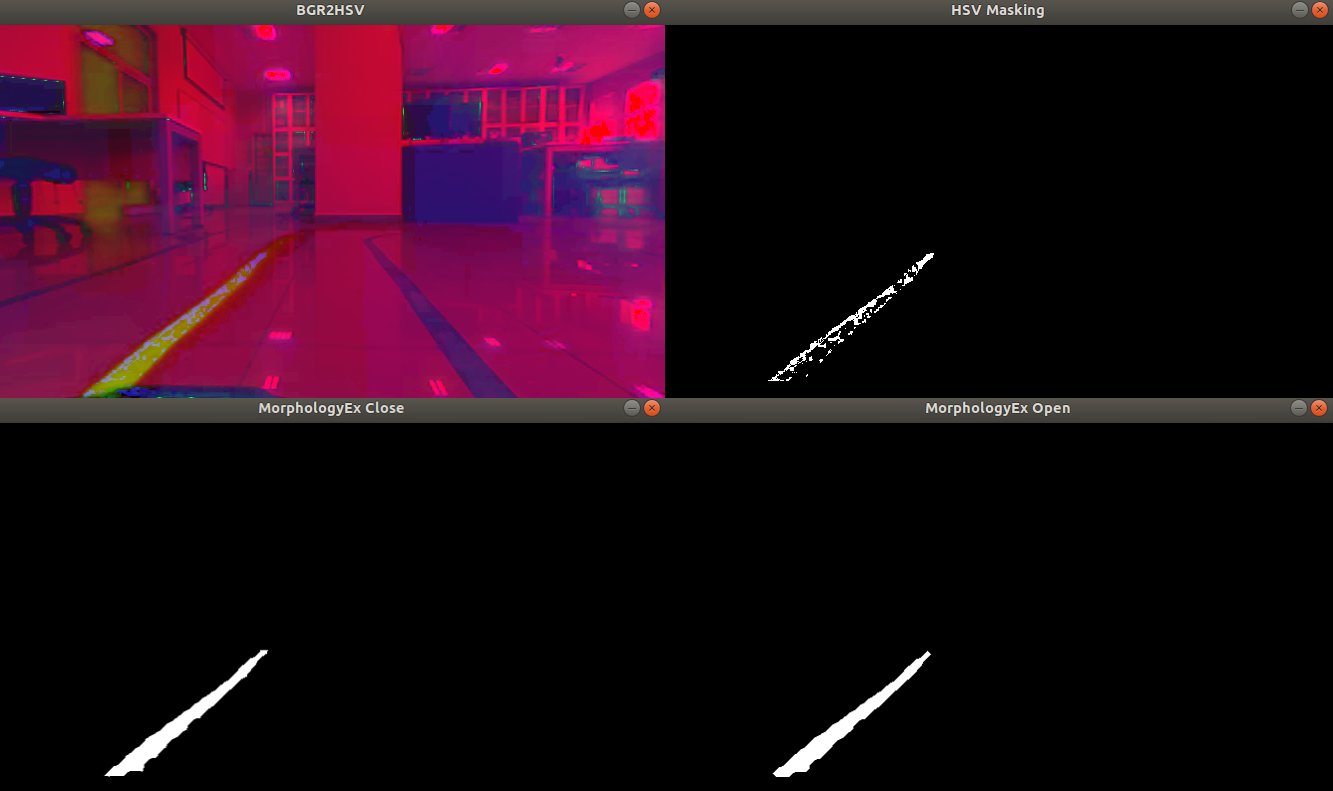}}
  \vspace{5mm}
  \caption{Example output of pre-processing image with stages}
  \vspace{5mm}
  \label{fig:pre_2}
\end{figure}

\begin{figure}[h!]   
  \centerline{\includegraphics[width=\linewidth]{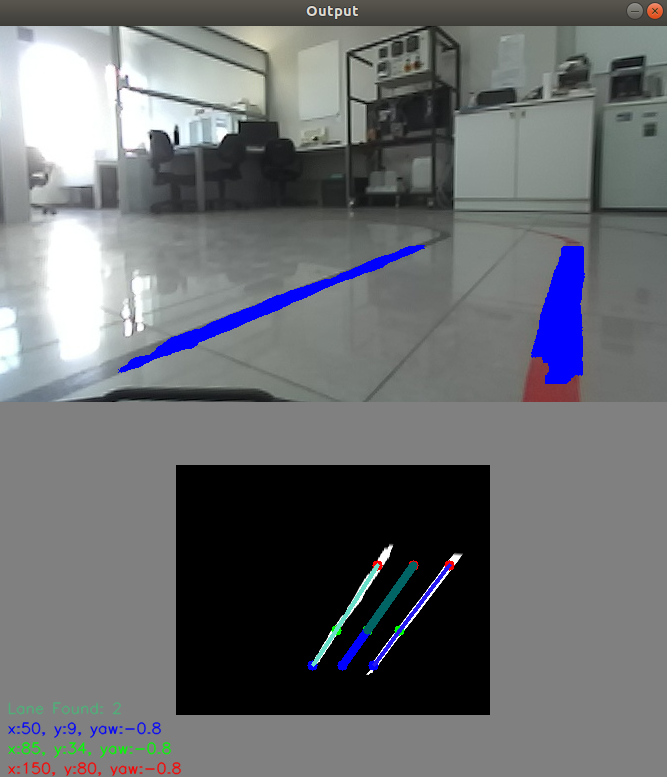}}
  \vspace{5mm}
  \caption{An example output of the lane detection algorithm}
  \vspace{5mm}
  \label{fig:detect_out}
\end{figure}

The second part of the algorithm takes the binary image that is the output of the pre-processing part as input. This part of the algorithm firstly takes the perspective transformation of the lane image to the Bird-Eye view. After taking transformation, the contour information of the image is extracted and applied some filtering again. After all of these processes, a basic second degree polynomial is fit for the lane. This is required because in some cases, only a small part of the lane is visible and the coordinates of the target point at the look ahead distance is required for path planning. Additionally, some coordinate transformations is applied to the target point and the point is passed as output of the lane detection algorithm. The flow of the entire algorithm can be seen in Figure \ref{fig:entire_flow} and an example output of the lane detection algorithm is also can be seen in Figure \ref{fig:detect_out}.

\clearpage

\begin{landscape}

\begin{figure}[h!]   
  \centerline{\includegraphics[width=\linewidth]{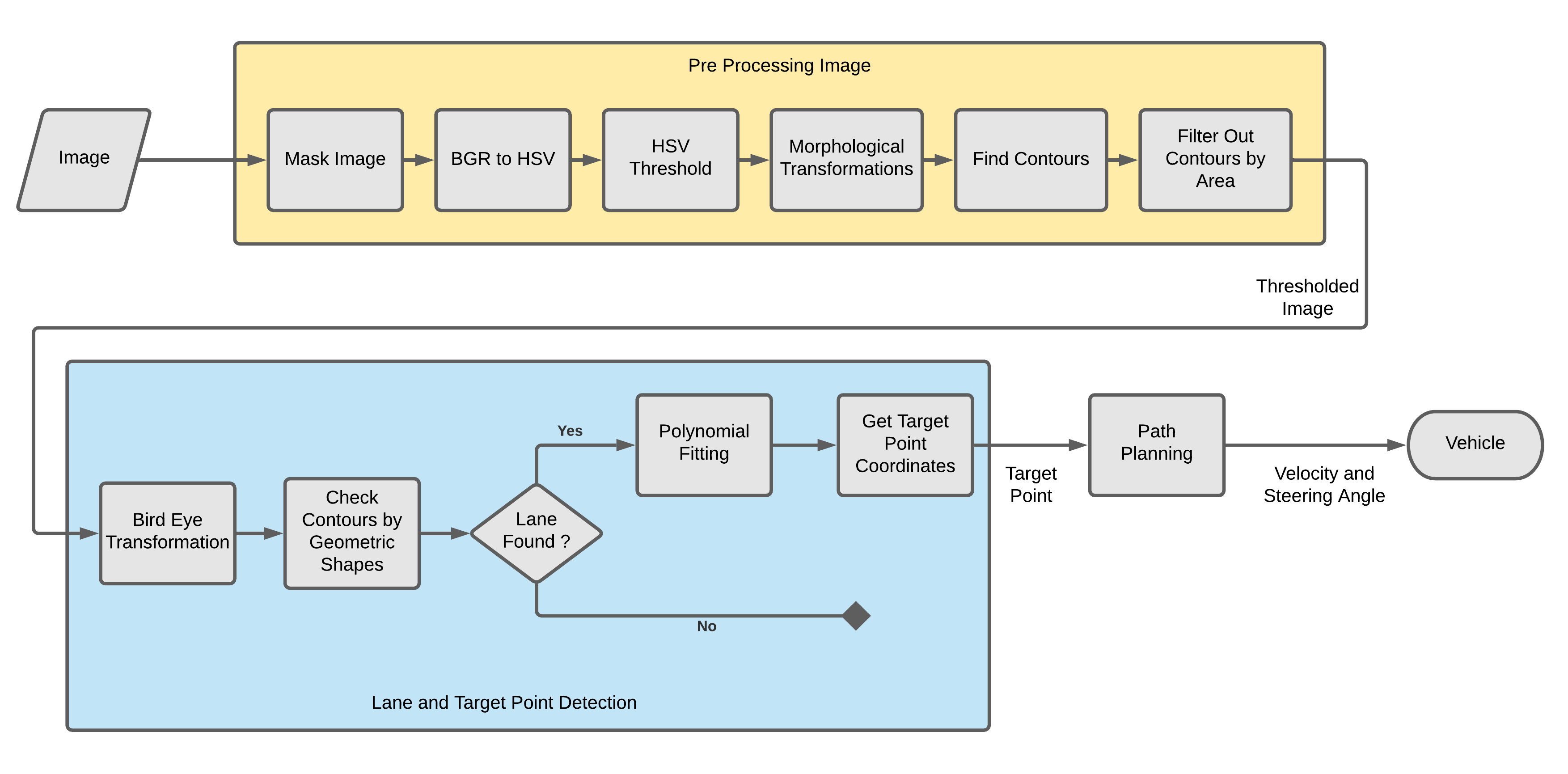}}
  \vspace{5mm}
  \caption{The flow of the entire lane detection algorithm}
  \label{fig:entire_flow}
  \vspace{5mm}
\end{figure}

\clearpage

\end{landscape}

\clearpage

\section{Testing and Results}

In this chapter, an overall qualitative assessment of the trajectory planning algorithms is given. This assessment relies on how successfully the vehicle followed the lanes, how many obstacles the vehicle pass through without collision and some special comments on the algorithm behavior. Since there is no global position data source in AI2S laboratory environment, an overall numeric error can not be calculated.

\begin{figure}[h!]   
  \centerline{\includegraphics[width=\linewidth]{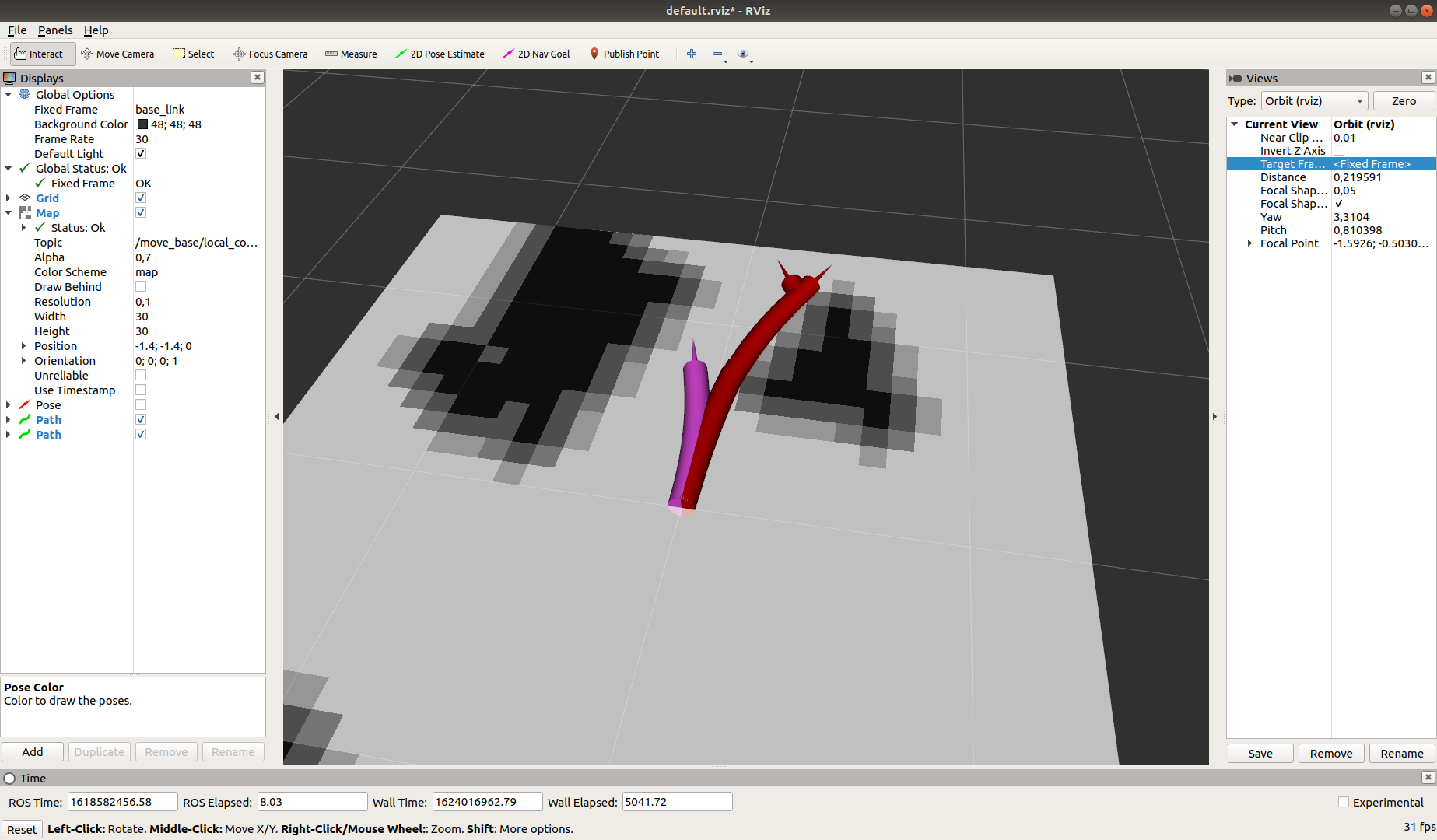}}
  \vspace{5mm}
  \caption{An example Rviz view with global plan (red) and local plan (purple)}
  \vspace{5mm}
  \label{fig:rviz}
\end{figure}

While testing the algorithms, it is tried to keep the environment same for all the algorithms. Nevertheless, some environmental circumstances like lighting condition can be changed. During the testing a Rviz is used which can be seen in Figure \ref{fig:rviz} for visualizing the vehicle condition from the vehicle's perspective, global plan and local plan that is manipulated by trajectory algorithm.

\begin{figure}[h!]   
  \centerline{\includegraphics[width=\linewidth, height=23cm]{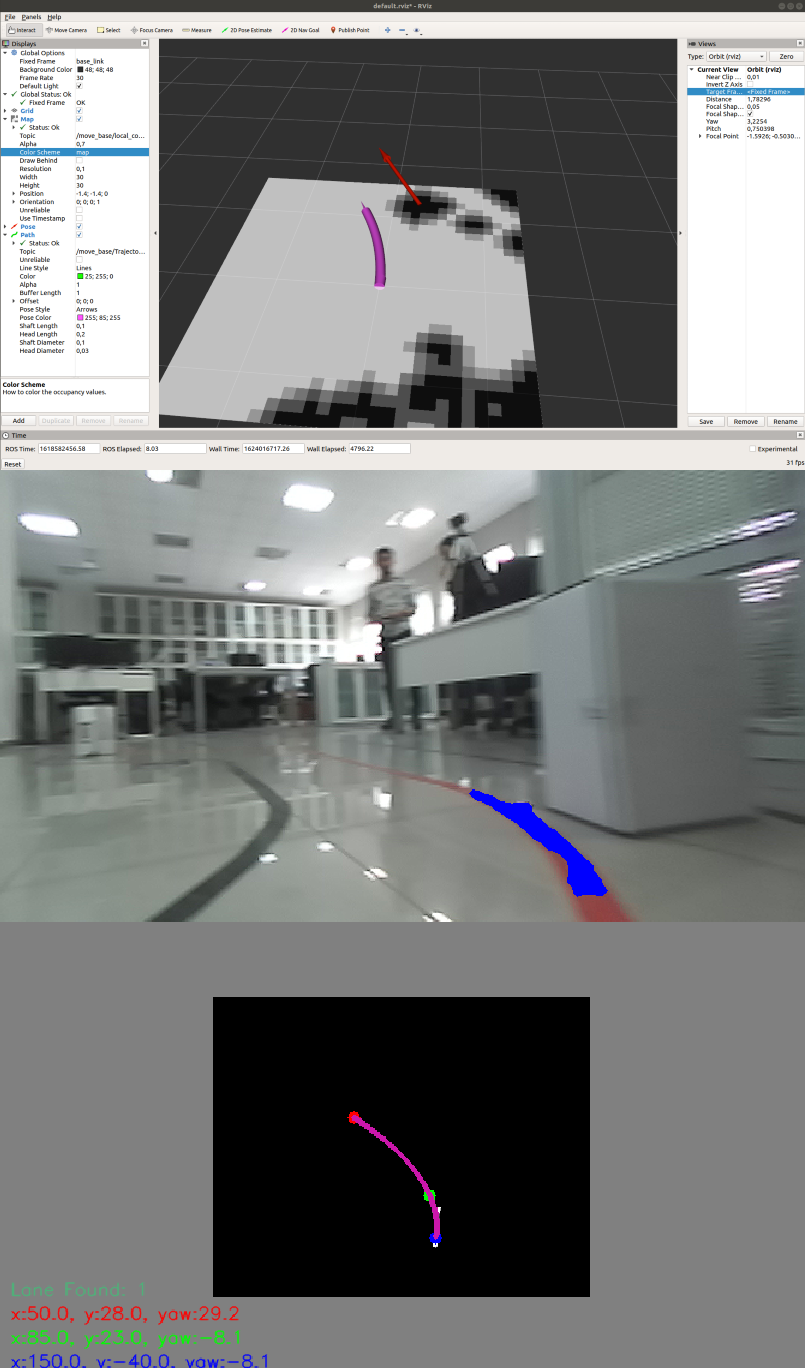}}
  \vspace{5mm}
  \caption{An example view that trajectory planning is running (red arrow is the output of the lane detection, the purple arrows are the planned trajectory)}
  \label{fig:manipulator}
\end{figure}

In addition, how the trajectory planning manipulates the target points which are determined by lane detection can be seen in Figure \ref{fig:manipulator}. At the top side of the figure, how the vehicle sees the environment can be seen and below the image from camera and lane detection algorithm can be seen. While the red arrow is the output of the lane detection algorithm, the purple arrows indicates the planned trajectory that will avoid the obstacle without leaving the road.

\clearpage

\clearpage

The final assessment about the trajectory planning algorithms can be inspected in Table \ref{tab:assessment}. But as a matter of fact, it should be said that this assessment is only according to the chosen scenery, it is not about which trajectory planning algorithm is better than the others. Besides, determining which algorithm is superior to others depends on the application scenery. As a conclusion, artificial potential field algorithm has better results according to the project requests and it is chosen to continue to the project with APF algorithm from now on. Since it provides reasonable performance with low cost, when considered possible future development of the project, the APF method is chosen.

\begin{table}[h!]
\centering
\begin{tabular}{@{}
>{\columncolor[HTML]{EFEFEF}}c 
>{\columncolor[HTML]{EFEFEF}}c 
>{\columncolor[HTML]{EFEFEF}}c @{}}
\toprule
\textbf{\begin{tabular}[c]{@{}c@{}}Trajectory Planning\\ Algorithm\end{tabular}} &
  \textbf{Obstacles Avoided} &
  \textbf{Comments} \\ \midrule
\begin{tabular}[c]{@{}c@{}}Dynamic Window\\ Approach\end{tabular} &
  5 out of 7 &
  \begin{tabular}[c]{@{}c@{}}DWA planner is easy to implement and tune. \\ Its performance was acceptable and stable. \\ Also, while avoiding obstacles, \\ it could stay in the course, but sometimes, \\ its effort was not enough to avoid obstacles. \\ Also, it gets stuck in complex situations\\  like narrow pass ways\end{tabular} \\ \midrule
\begin{tabular}[c]{@{}c@{}}Time-Elastic Band \\ Planner\end{tabular} &
  7 out of 7 &
  \begin{tabular}[c]{@{}c@{}}TEB planner is very successful to model the \\ vehicle, and it is suitable for complex \\ environments. But, it is very hard to tune \\ effectively and sometimes having so many\\  parameters to tune is turning into \\ a disadvantage instead of an advantage. \\ Also, since it heavily relies on \\ an optimization problem, computational \\ cost gets very high, especially \\ in complex environments.\end{tabular} \\ \midrule
\begin{tabular}[c]{@{}c@{}}Artificial Potential\\ Field\end{tabular} &
  5 out of 7 &
  \begin{tabular}[c]{@{}c@{}}APF is the easiest to implement and tune \\ by far. Thanks to its basic logic based \\ on simple math, it provides an acceptable \\ result with low cost. And also, \\ in line with the application scenery that is \\ selected for this project, it is not \\ too much deviates the vehicle from the road.\end{tabular} \\ \bottomrule
\end{tabular}
\caption{The assessment of the experiments of the trajectory generation algorithms}
\label{tab:assessment}
\end{table}
\chapter{Conclusion and Future Works}

Path planning algorithms are gaining more and more importance as autonomous vehicles become more widespread and important nowadays. There are many types of path planning applications used as the navigation unit of autonomous vehicles or for security purposes in semi-autonomous vehicles. In this project, which aims to implement and test a few of the path planning applications in real-time, a testing environment was first established and an image processing-based lane tracking algorithm was developed for this environment as reference signal to the vehicle. After the test environment was created, DWA, TEB and APF methods were implemented and tested separately. Since problems such as overtaking will be studied in the later parts of the project, it is of great importance to choose an algorithm that is predictable and low in cost. For this reason, it was deemed appropriate to continue the next stages of the project with APF. However, this does not mean that APF is better than others. However, all route planning applications have advantages and disadvantages over each other. 

In addition, in the next stages of the project, it is aimed to establish an end-to-end neural network structure and to carry out an end-to-end control with this structure. It is aimed to give sensor data such as camera and lidar as input to neural network and to obtain control signals such as vehicle speed and steering angle.

\bibliographystyle{itubib}
\bibliography{tez}

\ozgecmis{\vspace{10mm}

\newsavebox{\mysquare}
\savebox{\mysquare}{\textcolor{black}{\rule[2.3pt]{3.4pt}{3.4pt}}}

\setlength{\TPHorizModule}{10pt}
\setlength{\TPVertModule}{10pt}
\begin{textblock}{1}(40,10)
 \begin{figure}[p]
 \vspace{4cm}
 \rotatebox[origin=c]{180}{\includegraphics[scale=0.0035,keepaspectratio=true]{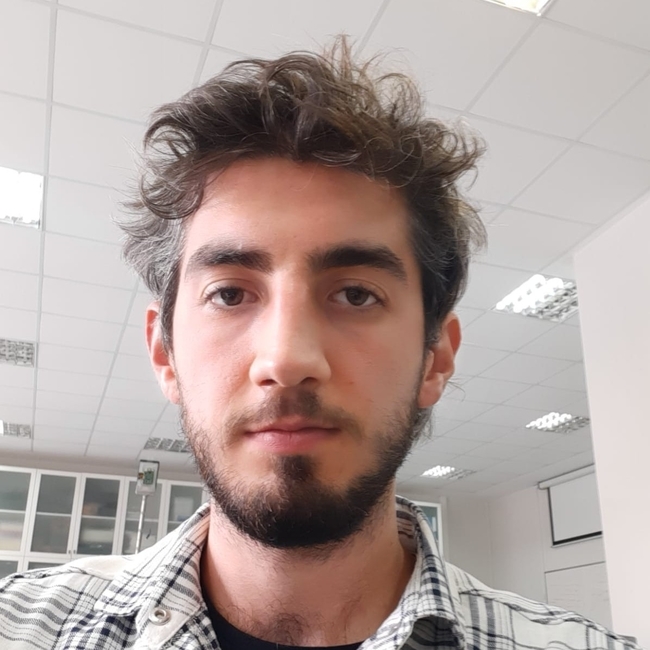}}
\end{figure}

\end{textblock}
\textbf{Name Surname:}  Oğuzhan Köse \\

\vspace{-3mm}
\textbf{Place and Date of Birth:}  Istanbul/Turkey, 24.09.1997 \\

\vspace{-3mm}
\textbf{E-Mail:} koseo16@itu.edu.tr\\

\textbf{EDUCATION:} 
\vspace{-3mm}
\begin{itemize}
  \item \textbf{B.Sc.:} 2021, Istanbul Technical University, Faculty of Electrical and Electronics Engineering, Control and Automation Engineering Department
\end{itemize}

\textbf{PROFESSIONAL EXPERIENCE AND REWARDS:}   
\vspace{-3mm}
\begin{itemize}
  \item 2020- Researcher at Istanbul Technical University Artificial Intelligence and Intelligent Systems (AI2S) Laboratory.
  \item 2019 Teknofest Fighting UAV Challenge, First Place in Fixed-Wing Category
  \item 2017 AUVSI Student Unmanned Aircraft Systems Competition, Best Rotary Wing Award, 4th place Overall Ranking
\end{itemize}
}

\end{document}